\documentclass[10pt,twocolumn,letterpaper]{article}

\usepackage[arxiv]{iccv}      %

\usepackage{microtype}
\usepackage{graphicx}
\usepackage{booktabs} %
\usepackage{xcolor}

\usepackage{amsmath, amsthm, amssymb}

\newcommand\daniel[1]{}
\newcommand\seb[1]{}
\newcommand\jj[1]{}

\newcommand\gelant{GELAN-t}
\newcommand\gelantransposed{GELAN-t$^T$}

\newcommand\gelanm{GELAN-m}
\newcommand\gelanmtransposed{GELAN-m$^T$}

\definecolor{iccvblue}{rgb}{0.21,0.49,0.74}
\usepackage[pagebackref,breaklinks,colorlinks,allcolors=iccvblue]{hyperref}

\newtheorem{definition}{Definition}

\def\paperID{14914} %
\def\confName{ICCV}
\def\confYear{2025}

\title{You Only Look Once at Anytime (AnytimeYOLO):\\ Analysis and Optimization of Early-Exits for Object-Detection}

\author{Daniel Kuhse\\
Technische Universität Dortmund\\
Germany\\
{\tt\small daniel.kuhse@tu-dortmund.de}
\and
Harun Teper\\
Technische Universität Dortmund\\
Germany\\
{\tt\small harun.teper@tu-dortmund.de}
\and
Sebastian Buschjäger\\
\small{
Lamarr Institute for Machine}\\
\small{
Learning and Artificial Intelligence}\\
Germany\\
{\tt\scriptsize sebastian.buschjaeger@tu-dortmund.de}
\and
Chien-Yao Wang\\
Academia Sinica\\
Taiwan\\
{\tt\small kinyiu@iis.sinica.edu.tw}
\and
Jian-Jia Chen\\
Technische Universität Dortmund\\
Germany\\
\small{
Lamarr Institute for Machine}\\
\small{
Learning and Artificial Intelligence}\\
Germany\\
{\tt\small
jian-jia.chen@tu-dortmund.de
}
}

\begin{document}
\maketitle
\begin{abstract}

We introduce AnytimeYOLO, a family of variants of the YOLO architecture that enables anytime object detection. Our AnytimeYOLO networks allow for interruptible inference, i.e., they provide a prediction at any point in time, a property desirable for safety-critical real-time applications. 

We present structured explorations to modify the YOLO architecture, enabling early termination to obtain intermediate results. We focus on providing fine-grained control through high granularity of available termination points. First, we formalize Anytime Models as a special class of prediction models that offer anytime predictions. Then, we discuss a novel transposed variant of the YOLO architecture, that changes the architecture to enable better early predictions and greater freedom for the order of processing stages. Finally, we propose two optimization algorithms that, given an anytime model, can be used to determine the optimal exit execution order and the optimal subset of early-exits to select for deployment in low-resource environments. We evaluate the anytime performance and trade-offs of design choices, proposing a new anytime quality metric for this purpose. In particular, we also discuss key challenges for anytime inference that currently make its deployment costly.

\end{abstract}

\section{Introduction}

A central challenge for algorithms in cyber-physical systems is the need to balance performance, timeliness, and resource utilization. In many applications, the time it takes to make a decision is as important as the quality of the decision, with a late result being sometimes as bad as a wrong result. Furthermore, from both cost and sustainability perspectives, the system resources should be used efficiently. 

\begin{figure}
    \centering
    \includegraphics[width=\columnwidth]{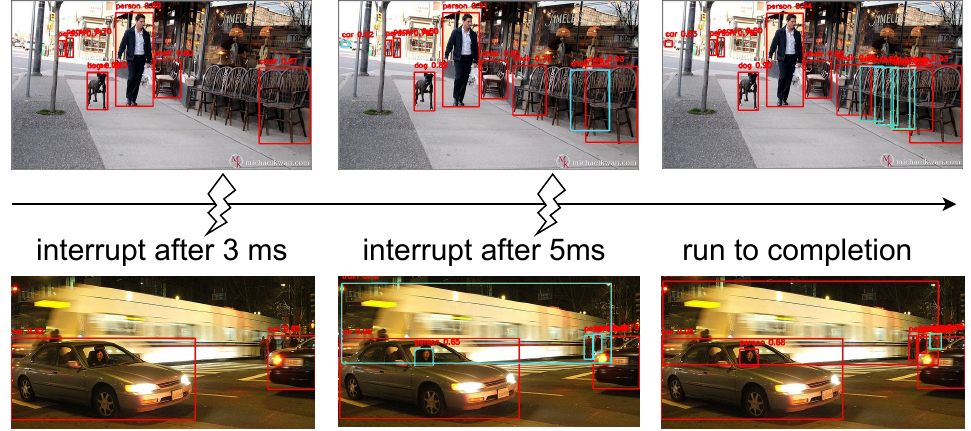}
    \caption{AnytimeYOLO on the MS COCO dataset~\cite{lin2014microsoft}. The model is interrupted at different points in time, providing a prediction. Longer runtime leads to better predictions.}
    \vspace{-3ex}
    \label{fig:anytime_inference}
\end{figure}

\emph{Anytime algorithms}, which can be \emph{interrupted} at any time during execution and still provide a valid result, c.f. \cite{DBLP:conf/ecai/TeijeH00,DBLP:journals/aim/Zilberstein96}, are a class of algorithms that address this challenge.  In contrast to traditional algorithms and machine learning models that must run to completion before providing a result, interruptible anytime algorithms and models enable the possibility to stop the execution \emph{during inference} and provide more flexibility for cyber-physical systems. An example of anytime object detection is given in Fig.~\ref{fig:anytime_inference}. For two images, model execution is interrupted at different points in time, respectively at 3~ms, 5~ms, and completing. One can see that the model successfully detects more objects the longer it runs, marked in light blue.  Another way to achieve a trade-off between quality and resource efficiency is using \emph{contract algorithms}, which specify a budget before execution. If a contract algorithm is interrupted before the budget is reached, it might fail to provide a valid result, whereas an anytime algorithm will provide a valid result at any point in time. Further discussions about the differences and applicability of anytime and contract algorithms can be found in \cite{DBLP:journals/aim/Zilberstein96}.

As a result,  for systems with resource constraints, e.g., cyber-physical systems or embedded systems, there has been a growing interest in anytime inference for object detection models, e.g.,~\cite{liu2020removing,liu2022self,kang2022dnn,soyyigit2024valo, soyyigit2022anytime, liu2021real, heo2020real, DBLP:conf/rtcsa/YaoHZSLLW0A20}. Most of them adopt the concept of \emph{early-exits} by adding early exit points to the corresponding neural network to enable anytime inference. They propose additional mechanisms to execute or use a neural network, e.g., prioritizing certain input data (by setting a certain focus area), scheduling the execution based on the expected gain of quality, and merging, scaling down, or reducing input images. 

To the best of our knowledge, for object detection, there is no anytime inference model with an in-depth exploration of \emph{how early-exits are added internally in the neural network}.
Computer vision models and especially YOLO networks have been proven invaluable as part of vision pipelines in cyber-physical systems, with applications ranging from shuttlecock detection in sports to vehicle and pedestrian detection for self-driving cars~\cite{DBLP:journals/mta/DiwanAT23}.

Our paper focuses on how to add anytime features to the YOLO models by studying how early-exits should be added internally in YOLO.
Our contributions are as follows:

\begin{itemize}
    \item We present the first formal adaptation of the notion of anytime algorithms to anytime models and derive a common evaluation metric from it in Section~\ref{sec:anytime}.
    \item We investigate the YOLOv9~\cite{DBLP:conf/eccv/WangYL24} model and introduce AnytimeYOLO, a novel architecture with early-exits for anytime inference, including a transposed variant that enhances early performance by reorganizing processing stages at a small cost to final accuracy in Section~\ref{sec:anytime_yolo}.
    \item We present optimization algorithms for anytime inference: determining the optimal execution order and selecting an optimal subset of exits to reduce memory and computation costs while maintaining performance in Section~\ref{sec:deployment_optimization}.
    \item We discuss deployment considerations for anytime inference, including execution with and without grace periods, and current challenges in deploying anytime models due to limited framework support in Section~\ref{sec:anytime_execution_deployment}.
    \item We evaluate our models on the MS COCO dataset~\cite{lin2014microsoft} and discuss the trade-offs of design choices in Section~\ref{sec:evaluation}.
\end{itemize}

\section{Related Work}
There is a rich literature on early-exits. Initially introduced by BranchyNet \cite{teerapittayanon2016branchynet}, early-exits add branches to the network that allow early termination. While introduced as a form of regularization, Stochastic Depth \cite{huang2016deep} can be used as a form of early-exiting, skipping layers \cite{elhoushi2024layer}. Another direction is routing-based, such as SelectiveNet \cite{geifman2019selectivenet}, where a small network decides which exit to take~\cite{DBLP:conf/cvpr/WuNKRDGF18, DBLP:conf/eccv/WangYDDG18}. A lot of research focuses on how to train early-exit networks \cite{laskaridis2021adaptive}, including techniques such as progressive shrinking \cite{cai2019once}. 

Vision transformers~\cite{dosovitskiy2020image, carion2020end} are  alternatives to convolutional neural networks for vision tasks. Due to being computationally expensive, early-exiting is a promising approach to make them more efficient \cite{bakhtiarnia2021multi,bakhtiarnia2022single,xu2023lgvit,yu2023dynamic}. 
Adadet~\cite{yang2023adadet} is an object detection network with early-exits, using earlier detection outputs to determine an uncertainty score to decide when to stop. It does not focus on how to add exits. The base model's performance is not competitive with recent YOLO versions, making direct comparison difficult, though its routing approach could be combined with our AnytimeYOLO.

DynamicDet \cite{lin2023dynamicdet} is an object detection model using a routing approach with two backbones (\emph{easy} and \emph{hard}), the second receiving both the image and the output of the first, and a routing network that decides whether to continue to the hard backbone. One possible backbone is YOLOv7. However, it is not suitable for anytime inference, as it has only two exits and focuses only on large variants.

Anytime object detection has been considered in the context of real-time systems, however, the focus of this research is on \textit{using} models with early-exits, developing heuristics to section the input, prioritize parts, and scheduling them accordingly \cite{liu2020removing,liu2022self,kang2022dnn,soyyigit2024valo, soyyigit2022anytime, liu2021real, heo2020real, DBLP:conf/rtcsa/YaoHZSLLW0A20}. Our structured exploration of adding early-exits to YOLO is orthogonal and could generally be combined with these methods.

A related, but distinct concept is deep supervision \cite{wang2015training, shen2019object}, which adds branches only during training to improve gradient flow. YOLOv7 \cite{DBLP:conf/cvpr/WangBL23} introduced deep supervision to YOLO, forming the first step of adding early-exits to YOLO.

\section{Towards Anytime Models}
\label{sec:anytime}
Informally, an anytime algorithm can be stopped at any time, yielding a valid result that improves with more runtime. More formally, the literature identifies several key properties for an anytime algorithm \cite{DBLP:conf/ecai/TeijeH00}. We highlight the four most important ones: (1) \emph{interruptability} --- the algorithm can be stopped at any time, providing a valid answer. (2) \emph{monotonicity} --- the answer should get better over time. (3) \emph{consistency} and \emph{diminishing returns} --- the improvements diminish over time, but are the same for each repetition of the algorithm on the same input. (4) \emph{measurable quality} --- the quality of the result can be determined precisely.

We adapt these definitions to our notion of \emph{anytime models} -- models that can be stopped at any time during execution and still provide a valid prediction. To do so, we consider a supervised learning setting with observations $x$ and labels  $y$ drawn from a joint distribution $x,y \sim \mathcal D \subset \mathcal{X} \times \mathcal Y$. Further, we assume a given quality function $q \colon \mathcal Y \times \mathcal Y \to \mathbb R$ to score prediction against labels. For example, in the case of object detection, it might be the mAP. We \emph{assume} the system can interrupt model execution at any time (e.g. in an emergency) and the model can give a prediction immediately (technical challenges discussed in Section~\ref{sec:execution_chunks}). Similar to anytime algorithms, our anytime models should be \emph{interruptible} by offering a valid prediction at any point in time, as well as \emph{monotone}, improving quality over time. Regarding \emph{consistency} and \emph{diminishing returns}, most machine learning models are typically deterministic during deployment, since usually no randomness is involved after training. They also converge over time against a fixed output that does not improve anymore (e.g. after complete execution). Hence, we combine the notion of \emph{consistency} and \emph{diminishing returns} into \emph{bounded returns}, meaning that the output converges to a stable prediction given enough time. We formalize this with the following definition.

\begin{definition}{(Anytime Model)}
Let $\mathcal D \subset \mathcal X \times \mathcal Y$ be a joint distribution and $q \colon \mathcal Y \times \mathcal Y \to \mathbb R$ a quality function. A model $f$ is an anytime model \underline{iff} it has the following properties
\begin{itemize}
    \item (interruptability) $f \in \mathcal F = \{f\colon \mathbb R_{\ge 0} \times \mathcal X \to \mathcal Y\}$
    \item (bounded return)
      \vspace{-1ex}
      \begin{align*}
        \forall x \in \mathcal X:& \exists T \in \mathbb R_{\ge 0}: \forall t \ge T: f(t, x) = f(T,x) \\   
        \forall x \in \mathcal X:& f(x) = \lim_{t \to \infty} f(t,x)
    \end{align*}
      \vspace{-4ex}
    \item (monotonicity) $\forall t_1, t_2 \in \mathbb R_{\ge 0}, t_1 \ge t_2: \mathbb E_{x,y\sim \mathcal D}\left[q(f(t_1, x),y)\right] \ge \mathbb E_{x,y \sim \mathcal D}\left[q(f(t_2, x),y)\right]$
\end{itemize}
\end{definition}

In traditional supervised learning, the main goal is to find a function $g\colon \mathcal X \to \mathcal Y$ for high-quality predictions given the input $x$. In contrast, an anytime function $f\colon \mathbb R_{\ge 0} \times \mathcal X \to \mathcal Y$ receives an additional input $t \in \mathbb R_{\ge 0}$, providing predictions at all time-points. We note that for any model $g$ there is a \emph{trivial anytime version}, by returning a default vector before $g(x)$ completes. Let $\theta \in\mathcal Y$ be a task-specific default vector (e.g. $\theta  = (0,\dots,0)$ can be the zero vector, or $\theta $ can be class priors based on the training data, etc.), then the trivial anytime version of $g$ is
$$
f(t, x) = \begin{cases} g(x) & \text{if $g(x)$ is done} \\ \theta  & \text{else}\end{cases}
$$

While the quality function $q$ gives the performance of a regular model directly, the performance of an anytime model is more nuanced as it progresses over time. For example, an anytime model might quickly reach acceptable performance but be outperformed by another model in the long run. 
To judge the performance of an anytime model, we now define its \emph{measurable quality} as follows.

\begin{definition}{(Quality of Anytime Model)}
\label{def:anytime_def}
Let $\mathcal D \subset \mathcal X \times \mathcal Y$ be a joint distribution, let $q \colon \mathcal Y \times \mathcal Y \to \mathbb R$ be a quality function and $w \colon \mathbb R_{\ge 0} \to \mathbb R$ be a weighting function. Further, let $f$ be an anytime model and let 
$$
T(x) = \min\{t|t \in \mathbb R_{\ge 0}, \forall \tau \ge t: f(\tau, x) = f(t,x)\}
$$
be the earliest point in time in which the output of $f$ does not change anymore. Then the quality of $f$ is given by
$$
Q = \mathbb E_{x,y\sim\mathcal D}\left[\frac{1}{T(x)}\int_0^{T(x)} q(f(t,x),y) w(t) dt \right] 
$$
\end{definition}

First, we introduce a weighting $w$ for applications where the quality at certain time points is more important. As an example, if the distribution of interrupts is known, we can appropriately weigh the quality of the model. If no such information is known, a uniform weighting $w(t)=1$ may be used.
Second, we normalize the quality regarding the total runtime until stabilization $T(x)$ given the model. When comparing multiple models, the same normalization factor $\frac{1}{T(x)}$ should be chosen for a fair comparison.

\begin{figure}
    \centering
    \includegraphics[width=0.7\columnwidth]{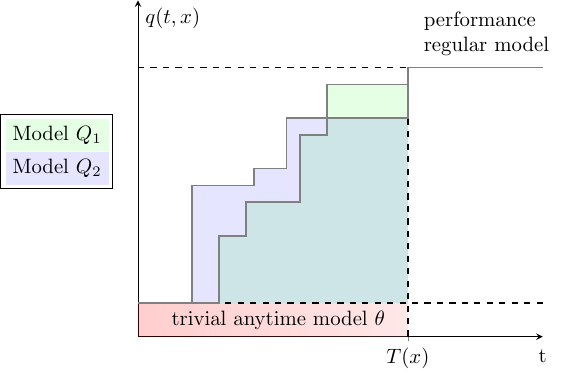}
    \caption{Anytime model quality, given observation $x$. The red area denotes the trivial anytime model quality, and the green area and blue areas are the quality of non-trivial anytime models. $T$ denotes the time point where the prediction quality stabilizes. The regular model produces predictions only after time $T$.}
    \vspace{-3ex}
    \label{fig:anytime_def}
\end{figure}

Figure \ref{fig:anytime_def} outlines these definitions: The models improve until they converge against a stable prediction at $T(x)$. The trivial anytime model (red area at the bottom) returns the default prediction $\theta$ until $T(x)$, yielding low anytime quality, while better anytime models (green or blue) continuously improve, with blue performing better at first until it is surpassed by green. 
We note that the interruptability requirement of the anytime model means the termination time is unknown beforehand. This is in contrast to other approaches where a time budget is given beforehand, though anytime models can be used for such budgeted execution as well.

\section{AnytimeYOLO}
\label{sec:anytime_yolo}

Given our definitions of anytime models, we now present a novel AnytimeYOLO version for real-time object detection. 
As mentioned earlier, we base our research on the YOLOv9 architecture~\cite{DBLP:conf/eccv/WangYL24}. Specifically, we focus on the smallest model available in the YOLOv9 family, \gelant{}, and the medium variant, \gelanm{}, as anytime inference is most relevant for resource-constrained embedded systems, though our approach applies to all YOLOv9 models.

Historically, YOLOv7 \cite{DBLP:conf/cvpr/WangBL23} added deep supervision to the YOLO architecture. Deep supervision involves adding another auxiliary detection head to the network in prior stages that is incorporated into the loss and trained alongside the entire network, intending to improve the performance of the final output. Moreover, to further improve the performance, YOLOv7 introduces a form of online distillation. During training, the auxiliary heads receive the predictions of the final outputs as guidance, whereas the final outputs receive the regular labels as usual. This form of self-knowledge distillation improves performance because earlier stages of the network are already guided towards meaningful representations that can be used for predictions. 

While deep supervision is not intended to enable anytime inference (auxiliary heads are normally removed for deployment), we repurpose deep supervision by adding multiple early-exits using the same scheme.

Fundamentally, as our goal is to offer good performance at any time, we aim to add as many early-exits as possible with as little (overall) performance degradation as possible, enabling high granularity for anytime inference.
A key part to consider for recent YOLO architectures is that they are multiscale object detection networks. That is, they perform object detection using feature maps at several different resolutions to more easily detect objects at different scales. As part of this, several \textit{feature pyramids} are used, where the features go through stages in which they get progressively smaller or larger~\cite{lin2017feature, liu2018path, bochkovskiy2020yolov4}. Such a feature pyramid consists of of two basic elements: two up/downscaling blocks and three GELAN blocks, which split the input, process it using several convolutions, concatenate it back together, and process it using a 1x1 transition convolution. Details can be found in the original YOLOv9 paper \cite{DBLP:conf/eccv/WangYL24} and we provide an overview in the supplemental material (Section~\ref{sec:suppl-implementation-details}).

\begin{figure*}
    \centering
    \begin{subfigure}[b]{\columnwidth}
        \centering
        \includegraphics[width=0.95\columnwidth]{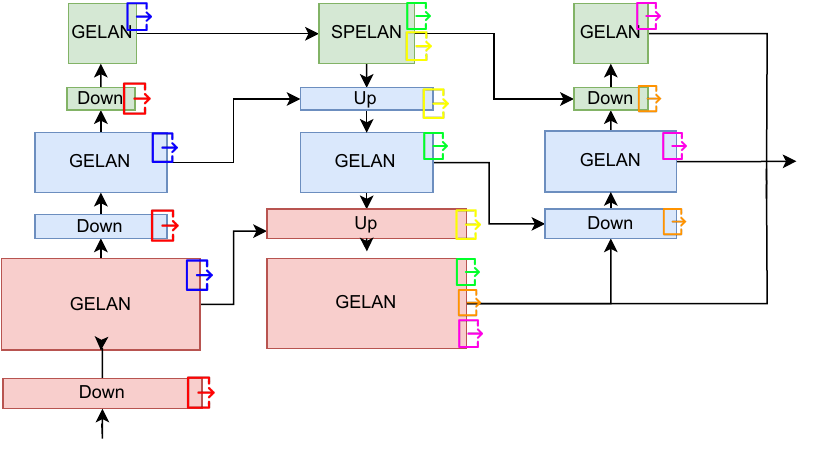}
        \caption{Architecture of \gelant{} and \gelanm}
        \label{fig:architecture-gelan-t}
    \end{subfigure}
    \begin{subfigure}[b]{\columnwidth}
        \centering
        \includegraphics[width=0.95\columnwidth]{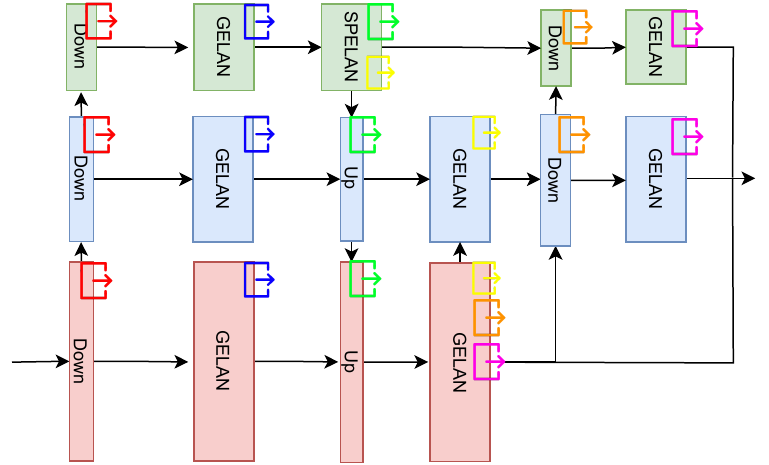}
        \caption{Architecture of \gelantransposed{} and \gelanmtransposed}
        \label{fig:architecture-transposed}
    \end{subfigure}
    \caption{Model Architectures: six medium scale layers, but only 4 large and 5 small layers (colored in blue, red, green respectively).} Sub-exits are marked using the exit symbol, colored depending on what exit they belong to for training. Red is the first exiit and violet for the last.
\end{figure*}

In the case of \gelant and \gelanm, three different feature map resolutions are used after the initial backbone. Fig. \ref{fig:architecture-gelan-t} shows the three feature pyramids with residual connections operating at three different resolutions. Detections using all three resolutions are combined to form the final detection $f(x)$:
\begin{equation*}
    f(x) = (f_s(g_s(x)), f_m(g_m(x), f_l(g_l(x)))),
\end{equation*}
\noindent where $g_s(x), g_m(x)$ and $g_l(x)$ are the feature maps output by the network stages operating at small, medium, and large resolutions and $f_s, f_m$ and $f_l$ the detection functions. For \gelant{} and \gelanm{} without early-exits, $g_s(x), g_m(x)$ and $g_l(x)$ are the predictions  of the last stages operating at small, medium, and large resolution. 

We therefore consider an early-exit to be comprised of what we call \emph{sub-exits}. Sub-exits themselves can be placed at the end of every stage (e.g. each GELAN block or any block derived from it) or after a down/upscaling block. That is, as can be seen in Fig.~\ref{fig:architecture-gelan-t}, marked in the diagram using the exit symbols, there are up to 15 meaningful sub-exits possible. For training purposes, these are matched with others of the same feature pyramid (indicated by the color of the exit symbol), however during inference, it is possible to freely match the sub-exits with each other. As an example, if one is at the end of the first stage of the last feature pyramid, one can then use an exit that uses two other sub-exits of the prior feature pyramid. For an anytime inference scheme, the idea is then to use the last sub-exit reached for each scale. While it is possible to have an early-exit not using all scales, this leads to drastic performance degradation. As part of our evaluation, we consider both variants where sub-exits are placed at every single location and only at select spots.

\subsection{Transposed architecture}
\label{sec:transposed}

In the pyramidal architecture, an early-exit can only use features from all resolutions at the end of the third stage after the backbone, i.e. the input data is always processed sequentially for every resolution before any prediction becomes available. This corresponds to an early-exit after about a third of the network has been executed. Without using all resolutions, the early-exit quality is comparatively low. Interestingly, the biggest quality gains are not after the comparatively slow GELAN blocks which are made up of several convolutions, but after the fast down/upscaling blocks which only feature an average pooling layer and a single convolution besides the down/upscaling operation. We shortly give measurements in our supplemental materials in Section~\ref{sec:suppl-full-evaluation}.

To improve its anytime capabilities, we propose a change to the overall architecture of the network: Instead of processing the image at three different resolutions sequentially, we aim to modify the network so that the resolutions are essentially parallel branches. This change allows not only the much earlier use of feature maps at all resolutions but also some degree of freedom for choosing the order in which the resolutions should be processed.

For this purpose, we metaphorically \textit{transpose} the stages, so that the stages are not dependent on the fully processed feature maps of the previous stage, but rather only the beginning of the stage. This transposed architecture, which we call \gelantransposed{} for the tiny variant or \gelanmtransposed{} for the medium variant, is shown in Fig. \ref{fig:architecture-transposed}. The first exit, using the output of all three stages after the backbone, then only requires the execution of the downscaling part of each stage.

\subsection{Training AnytimeYOLO}
\label{sec:training}

We modify YOLOv9's distillation-based loss and detection head implementation to support an arbitrary number of early-exits. As we can place one sub-exit after executing a block, the maximal number of sub-exits we can place is the number of blocks. As shown in Figure~\ref{fig:architecture-transposed}, \gelant{}, \gelantransposed{}, \gelanm{} and \gelanmtransposed can fit up to $15$ sub-exits. After an early-exit, to derive the distillation-based loss, all three resolutions must be considered. Therefore, the sub-exits of the three resolutions must be paired, leading to the fact that the number of sub-exits is a multiple of the number of final resolutions used, i.e., three for \gelant{} and \gelanm{}. As the model architecture in Figure~\ref{fig:architecture-transposed} does not have the same blocks per resolution, i.e., four, five, and six blocks for the large, small, and medium resolutions (red, green, and blue in Fig. \ref{fig:architecture-transposed}), respectively, we have to add redundant sub-exits (at the last GELAN blocks) to meet the requirement described above. As a result, using all medium-scale sub-exits requires training six pairs of three sub-exits, i.e., 18 sub-exits by introducing three redundant sub-exits.

The two most common strategies for training early-exit networks are \emph{joint} optimization, e.g.,~\cite{laskaridis2021adaptive}, and 
\emph{probabilistic} optimization, e.g.,~\cite{cai2019once}, that samples exits. In joint optimization, all early-exits are trained at each training step, with the loss being a sum of the losses of each exit, possibly with the loss being weighed differently for each exit~\cite{laskaridis2021adaptive}. 
Probabilistic approaches sample a subset of exits at each training step~\cite{cai2019once}, either using a fixed set of early-exits or by progressively sampling more exits during training. One such approach is progressive shrinking~\cite{cai2019once}, where early-exits that process larger subnetworks are added first. 

Our final consideration to train AnytimeYOLO is \emph{pre-training}.  Progressive shrinking~\cite{cai2019once} trains the full network first and then adds early-exits. It reuses non-anytime network training and reduces computational costs. However, since the network initially optimizes only for the final exit, the anytime version might favor the final exit, resulting in worse performance when the anytime inference exits much earlier.

\section{Anytime Deployment Optimizations}
\label{sec:deployment_optimization}

Early-exits can increase network memory consumption and computational cost during inference due to additional prediction heads. While traditional deep learning architectures execute \emph{all} layers in an order optimized for speed (e.g., minimizing memory accesses), anytime models require \emph{executing the layers with care} as early-exits may depend on specific layers. As a result, the selection of early-exits and the scheduling of layers are critical for the deployment of anytime models. A non-optimized execution order of the layers can degrade the predictive performance.

We handle the selection of early-exits and the scheduling by graph theory, optimizing for anytime quality as given by Definition~\ref{def:anytime_def}. We model the execution of the anytime model as a directed graph $G = (V, E)$. Given a set of network layers (or blocks) $L$, let a node $v$ in $V \subseteq 2^{L}$ represent a set of executed layers. The empty set $\emptyset$ means that no layer has been executed, while the full set $L$ means that the entire network has been executed. Let the set of edges $E$ be determined by the dependencies of layers, such that a path in $G$ from $\emptyset$ to $L$ is equivalent to a valid execution order $o = (o_1, ..., o_n)$ for the network, with $o_i \in V$. That is, an execution order is valid if it complies with the execution dependencies of the neural network and the layers of $o_i$ are all in $o_j$ when $i \leq j$.

We aim to label the edges such that the cost of a path from start to finish is equivalent to the anytime quality. We achieve this by splitting up the integral into a part for each element in an order. Given the anytime model executed in the order of $o$ as $f_o$, Definition~\ref{def:anytime_def} can be altered to define the order-dependent anytime quality $Q_o$ as:
\begin{equation*}
    Q_o = \mathbb{E}_{x,y\sim\mathcal D}\left[\int_0^T q(f_o(t,x),y) w(t) dt \right]
\end{equation*}
Since the execution time $T$ of each order is the same, we leave out the normalizing factor $\frac{1}{T}$.

We now split the integral into a sum over the elements in an order $o$. Let $c(v)$ be the sum of the average execution time of all layers in $v$. The overall anytime quality $Q_o$ can then be written as:
\begin{equation*}
\label{eq:expanded_performance}
    Q_o = \sum_{i=1}^{n-1} \mathbb{E}_{x,y\sim\mathcal D} \left[\int_{c(o_i)}^{c(o_{i+1})}q(f_o(t,x),y) w(t) dt\right] 
\end{equation*}

We now label the edges of $E$ with the corresponding parts of the sum. With $t$ in the interval $[c(v_i),c(v_{i+1})]$, we know that $q(f_o(t,x), y)$ is independent of the order $o$ and is solely determined by $v_i$, i.e., the expected quality of the exit reached after executing $v_i$, denoted as $q_{v_i}$. Let the weight $\rho{(v_i, v_j)}$ of an edge $(v_i, v_j)$ in $E$ of the graph be:
\begin{equation*}
    \rho{(v_i, v_j)} = \mathbb{E}_{x,y\sim\mathcal D} \left[ \int_{c(v_i)}^{c(v_j)} q_{v_{i+1}}(x,y)  w(t) dt \right]
\end{equation*}

Therefore, a directed path in $G$ corresponds to an execution order $o$ with the sum of the edge weights of the path equal to its performance $Q_o$.  The problem of finding the optimal execution order is then equivalent to finding the \textit{longest} path. We consider that the original neural network is a directed acyclic graph (DAG). Therefore, every layer is executed at most once. Our construction procedure of $G$ ensures that it is also a DAG. The \textit{longest} path problem of a DAG can be handled by negating the edge weights and then adopting the well-known Bellman-Ford algorithm~\cite{ford1962flows, bellman1958routing} for the single-source shortest path problem~\cite{sedgewick2011algorithms}.

For the problem of selecting exits, we are given a target amount of exits $k$ (not including the start point). We take the previously defined graph and only consider the nodes where new exits are available. To get the best $k$ exits, we terminate the algorithm early after $k$ iterations, as this yields the longest path with $k$ edges from the start  to the end node.

As mentioned in Section~\ref{sec:anytime_yolo}, AnytimeYOLO has the notion of sub-exits, where each exit is comprised of three separate exits. 
For our method to select sub-exits, we need to further restrict the set of edges so that a node can only be reached from another if they only differ in one sub-exit.
Each edge of the longest path then corresponds to a single sub-exit of the selected set of exits.

\section{Anytime Execution Deployment}
\label{sec:anytime_execution_deployment}

In this section, we discuss how a model with early-exits can be executed in an anytime manner.

\subsection{Soft and hard anytime inference}
\label{sec:soft_hard_anytime}   
We distinguish between two types of anytime inference, depending on whether a grace period to finalize the computation is given (namely, \emph{soft anytime inference}) or not given (namely, \emph{hard anytime inference}) after the termination signal has been sent. Concretely, the difference is whether the early-exit is allowed to be executed after the termination signal, incurring a delay equal to its execution time and the time it takes to transfer the output to the user. 

Hard anytime inference is more challenging as it requires intermediate outputs to be immediately available. If an early-exit's result is to be used, then it must be computed before the algorithm is terminated. This means early-exits must be executed whenever they are reached. Many computed intermediate results are then discarded, and replaced by a result further down the network. It is possible to use earlier predictions to enhance later predictions ensemble style or speed up computations in autoregressive transformers \cite{wolczyk2021zero,elhoushi2024layer}, but applicability depends heavily on the domain. There is therefore a trade-off between the number of early-exits and the computational cost of the network. With many exits, this can pose a significant cost, potentially eclipsing the cost of the network itself. See the supplementary material in~Section~\ref{sec:suppl-deployment-measurements}.

In the soft anytime case, the exit is executed only once the termination signal is sent. The number of early-exits therefore has no impact on execution time, but one must wait for the exit's execution time, which can be long.

\subsection{Anytime Execution Hurdles}
\label{sec:execution_chunks}
We briefly discuss how networks with early-exits can be executed in an anytime manner and the hurdles currently faced. Anytime inference requires the ability to interrupt execution and return a previously computed result. CPU inference is straightforward as we can simply execute one layer after another and store results in a predefined memory block so that whenever we are interrupted this memory block contains the most recent results. %

However, deployment of AnytimeYOLO on GPUs is currently more challenging. CUDA kernels are, by design, executed asynchronously. Hence, queued layers (i.e. queued CUDA kernels) cannot be canceled or interrupted without support by CUDA itself. Queueing up many layers asynchronously is typically more efficient, motivating the asynchronous execution of the entire model at once. Furthermore, most inferencing frameworks such as ONNXRuntime, TensorRT, or Tensorflow Lite typically perform network optimization on the entire network (e.g. layer fusion) and execute the entire model as a whole. 

One possibility to implement anytime support is to force the GPU kernels to check a global \texttt{interrupt} flag to return without doing work. However, this requires extensive modifications of the adopted deployment framework and may be error-prone. We seek a systematic approach by splitting the network into chunks that can be conditionally executed. Although this still requires regular synchronization on the host and incurs non-negligible overheads, no modification of the adopted framework is needed.

As a result, on GPUs, if the complete model is executed, our anytime inference is slower. The impact of anytime inference by utilizing TensorRT and TorchScript can be found in our supplementary material in~Section~\ref{sec:suppl-deployment-measurements}. Our anytime inference offers the possibility to interrupt the inference anytime and the overhead can be improved if there is any interrupt and cancellation feature supported by GPU vendors.

We note that these hurdles also apply to other AI accelerators like Neural Processing Units (NPUs), as they similarly execute operations asynchronously and are typically used with frameworks that optimize the entire network as a unit.

\section{Experimental Evaluation}
\label{sec:evaluation}

\begin{table}
    \centering
    \scalebox{0.7}{
    \begin{tabular}{l|p{1cm}|p{0.7cm}|p{0.9cm}|p{1.2cm}|p{1cm}|p{1.1cm}|p{1.6cm}}
    Name & Pretrain & Exits & $Q_{{AP}_{50}}$ & 
    $Q_{AP_{50}SE}$ & 
    AP$_{50}$ & 
    $\max \Delta$ & 
    MSize \\ \hline \hline
    YOLO9-m & - & - & - & - & 68.1 & 9.92 ms & 38.84 MB \\

    \gelanm & - & - & - & - & 67.62 & 9.98 ms & 38.84 MB \\

    \gelanmtransposed & - & - & - & - & 66.49 & 11.2 ms & 60.48 MB \\

    \hline
    
    \gelanm & 0 & 9 & 45.8 & 11.15 & 66.59 & 2.23 ms & 57.84 MB \\
    \gelanm & 0 & 15 & \textbf{47.07} & 10.41 & 67.29 & 2.23 ms & 84.84 MB \\
    \gelanm & 200 & 9 & 45.48 & 11.3 & \textbf{67.49} & 2.95 ms & 57.84 MB \\
    \gelanm & 200 & 15 & 46.62 & 10.63 & 67.42 & 2.23 ms & 84.84 MB \\
    \gelanmtransposed & 0 & 9 & 44.55 & 8.61 & 65.38 & 3.25 ms & 64.54 MB \\
    \gelanmtransposed & 0 & 15 & 46.3 & \textbf{8.31} & 65.69 & \textbf{1.62 ms} & 92.04 MB \\
    \gelanmtransposed & 200 & 9 & 43.26 & 9.14 & 65.86 & 4.44 ms & 64.54 MB \\
    \gelanmtransposed & 200 & 15 & 45.27 & 8.74 & 65.45 & \textbf{1.62 ms} & 92.04 MB \\
    \end{tabular}}
    \caption{Performance Comparison Table}
    \label{tab:performance}
\end{table}

We evaluate the performance of the medium variant of our proposed AnytimeYOLO model on the MS COCO benchmark \cite{lin2014microsoft} using standard train-test splits, following YOLOv9's protocol \cite{DBLP:conf/eccv/WangYL24}. Supplementary Section~\ref{sec:suppl-full-evaluation} provides full evaluations for both tiny and medium variants.

We omitted the PGI training technique from \cite{DBLP:conf/eccv/WangYL24} due to no direct compatibility with early-exits, resulting in our baseline \gelanm{} performing slightly below YOLOv9-m. We trained all models for 500 epochs on a single NVIDIA A100 GPU with batch size 32, taking~7 days per run, using the same hyperparameters as YOLOv9.

During evaluation, we focus on the following metrics:
\begin{itemize}
    \item \textbf{Non-anytime quality:} We evaluate the final exit's performance, i.e. the network's performance without early-exits. We use mean average precision (mAP) at an IoU of 0.5 as the evaluation metric, since it is standard in object detection, denoted AP$_{50}$. An evaluation for AP$_{50:90}$ is provided in the supplement materials, Section~\ref{sec:suppl-full-evaluation}.
    \item \textbf{Anytime quality:} Following our anytime quality in Definition~\ref{def:anytime_def}, we use the exit's mAP$_{50}$ reached as the quality function, with the integral of the mAP$_{50}$ curve as the primary metric $Q_{AP_{50}}$. We weigh everything equally for time. To weigh a relative performance increase more heavily, we also consider the squared error between the achieved performance and the non-anytime baseline as a secondary evaluation metric $Q_{AP_{50}SE}$.
    \item \textbf{Model Size:} We report the model size (MSize) to show the memory usage of weights.

    \item \textbf{Anytime Granularity:} We report the largest time step between exits, $\max \Delta$, to show the granularity of inference of AnytimeYOLO. For the reported time we use an A100, inference with batch size 1, measuring layers individually.
\end{itemize}

Preliminary experiments (detailed in supplementary Section~\ref{sec:suppl-preliminary-experiments}) informed our experimental design with several key findings: 1.) There is no clear evidence showing that more sub-exits are always better, leading to an exploration of selecting $k=9, 15$ sub-exits, 2.)  Applying 400 epochs for pre-training leads to worse results than 200 epochs in our preliminary experiments, leading us to choose either no pre-training or 200 epochs of pre-training, and 3.) We discuss the possibility of utilizing joint optimization and exit sampling in Section~\ref{sec:training} for training, finding the difference between them negligible, thus we only perform exit sampling.

As baseline, we train a variant without additional early-exits for both the \gelanm{}~and~\gelanmtransposed{}~architectures. For anytime performance evaluation of transposed models, we use the optimal path method from Section~\ref{sec:deployment_optimization}. For evaluating anytime performance, we use the soft anytime inference scheme, excluding detection head costs for unused exits.

In the experiment in Section~\ref{sec:performance}, we first explore possible design options in a structured manner. We then evaluate the performance of our proposed path selection algorithm in the experiment in Section~\ref{sec:path_selection_performance}. 

\subsection{Performance of AnytimeYOLO}
\label{sec:performance}

\begin{figure}
    \centering
    \includegraphics[width=0.65\columnwidth]{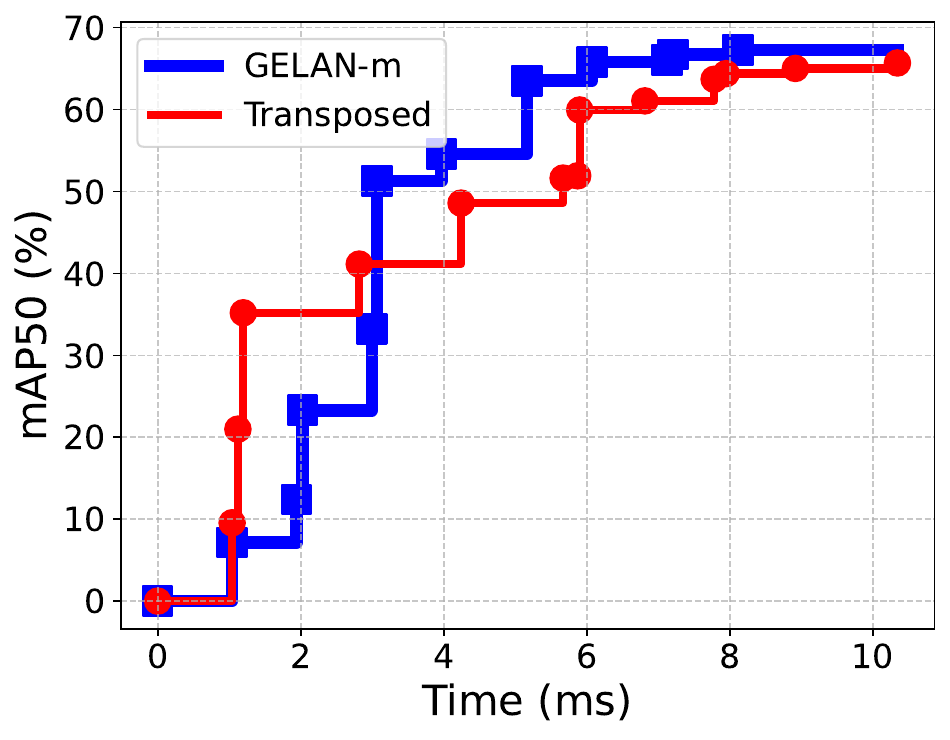}
    \caption{Quality of \gelanm{} and \gelanmtransposed{} with six trained exit, no pre-training.}
    \label{fig:performance-transposed-architecture}
\end{figure}

The quality of the different anytime models is shown in Table~\ref{tab:performance}. 
For completeness, we also include the non-anytime versions of YOLOv9-m,~\gelanm{}, and~\gelanmtransposed{} in the first three rows of Table~\ref{tab:performance}.

No model dominates. For example, if the target is the maximum anytime quality, the variant listed in row 9, \gelanmtransposed{} with 15 sub-exits and no pre-training, is the best, while if the target is maximum final precision, the variant listed in row 7, \gelanm{} with 15 sub-exits and 200 epochs pre-training, is the best.

Trade-offs are therefore necessary, with the best choice depending on the objectives. We observe several trends from Table~\ref{tab:performance} that should inform the choice of model: In comparison to \gelanm, our proposed \gelanmtransposed achieves better anytime quality with substantially better squared error anytime quality, but sacrifices final performance, most pronounced in the case of no pre-training. Pre-training improves final exit precision, but at the cost of some anytime quality in the 15 sub-exit cases. In the 9 sub-exit cases, anytime quality increases by a very small amount through pre-training in one case. For \gelanmtransposed{}, increased exit-count improves anytime quality, but worsens final exit quality. For \gelanm{}, 15 sub-exits achieve both superior anytime and final quality. Maximum time between exits is moderately reduced by a higher exit count. The amount of exits is the determining factor for model size.

\begin{figure}
    \centering
    \includegraphics[width=0.7\columnwidth]{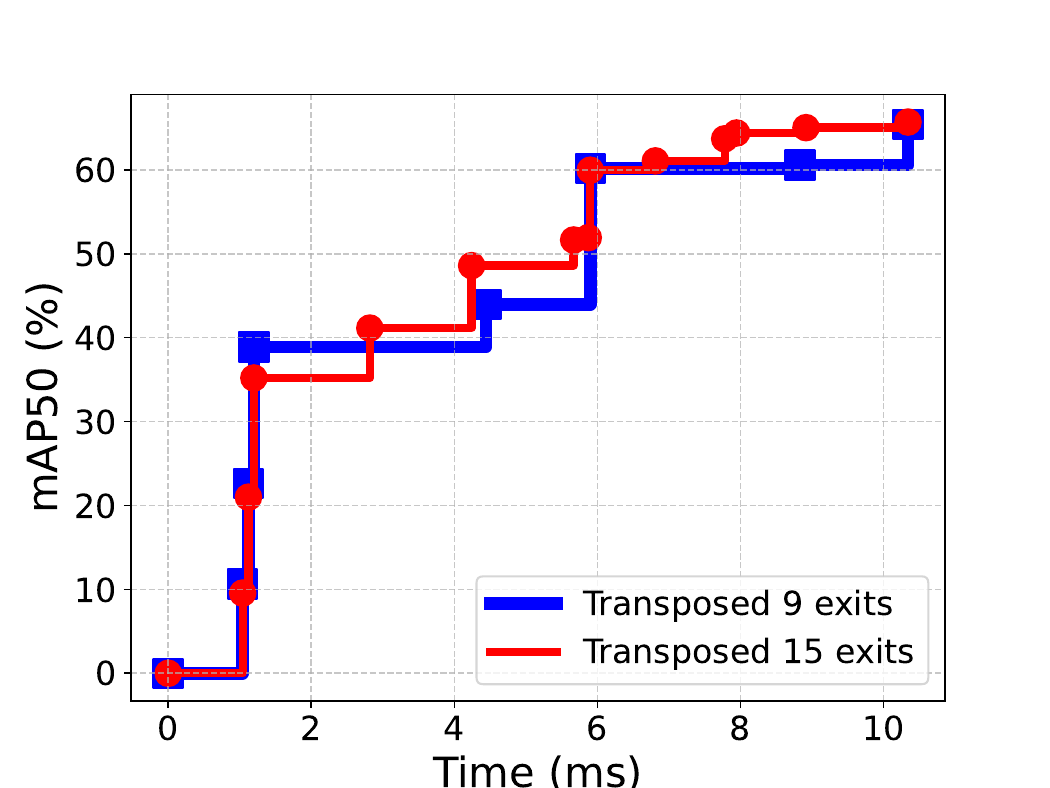}
    \caption{Quality of \gelanmtransposed{} with three and six trained exits, no pre-training.}
    \label{fig:performance-exit-count}
\end{figure}

Two anytime quality curve plots illustrate these results. The first, Fig.~\ref{fig:performance-transposed-architecture}, shows that \gelanmtransposed{} with 15 sub-exits compared to \gelanm{} with 15 sub-exits without pre-training has substantially better anytime quality early on, but falls off at the end. Fig.~\ref{fig:performance-exit-count} compares two variants of \gelanmtransposed{} without pre-training and different amounts of exits. With 9 sub-exits, the exit quality is slightly higher, but with less granularity, leading to slightly lower anytime performance.

\begin{figure}
    \centering
    \includegraphics[width=0.68\columnwidth]{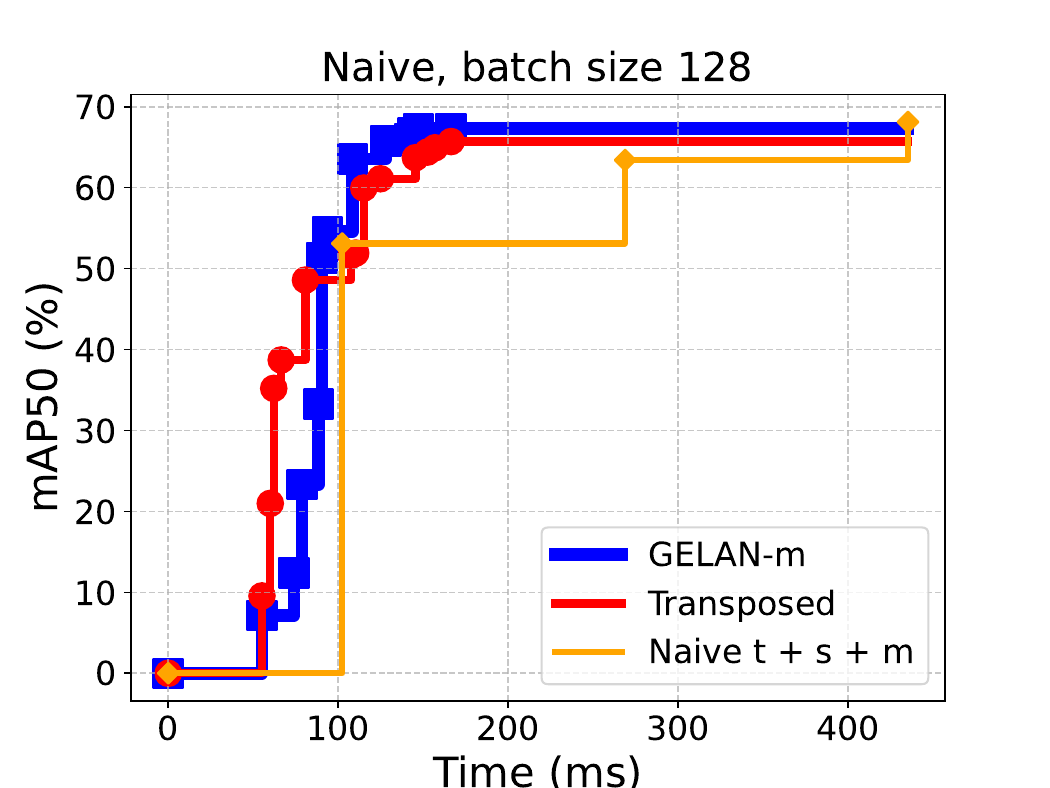}
    \caption{Comparison of AnytimeYOLO to running YOLOv9-t, YOLOv9-s, and YOLOv9-m sequentially, batch size 128.}
    \label{fig:sequential}
\end{figure}

A na\"ive alternative to anytime models is to simply run multiple models of different sizes sequentially. For this purpose, we compare our anytime models to running YOLOv9-t, YOLOv9-s, and YOLOv9-m sequentially. We evaluated using batch sizes of 1, 2, 4, 8, 16, 32, 64, and 128, which are fully reported in the supplementary materials, Section~\ref{sec:suppl-full-evaluation}. The anytime version processes the images of a batch in parallel, the first block of the model is run on all images, then the second block, and so on. Here, we report using a batch size of 128 to simulate a more realistic scenario, where the medium model takes significantly longer to process. The results are shown in Fig.~\ref{fig:sequential}. Anytime models prove more efficient, reaching equal performance in half the time with finer granularity.

\subsection{Path Selection Performance}
\label{sec:path_selection_performance}
The algorithm described in Section~\ref{sec:deployment_optimization} is optimal under the given assumptions. We compare it to two na\"ive heuristics, based on a greedy approach: \textit{greedy time}, choosing the next exit reachable in the shortest time, and \textit{greedy performance}, choosing the next exit with the best performance.

\begin{figure}
    \centering
    \includegraphics[width=0.7\columnwidth]{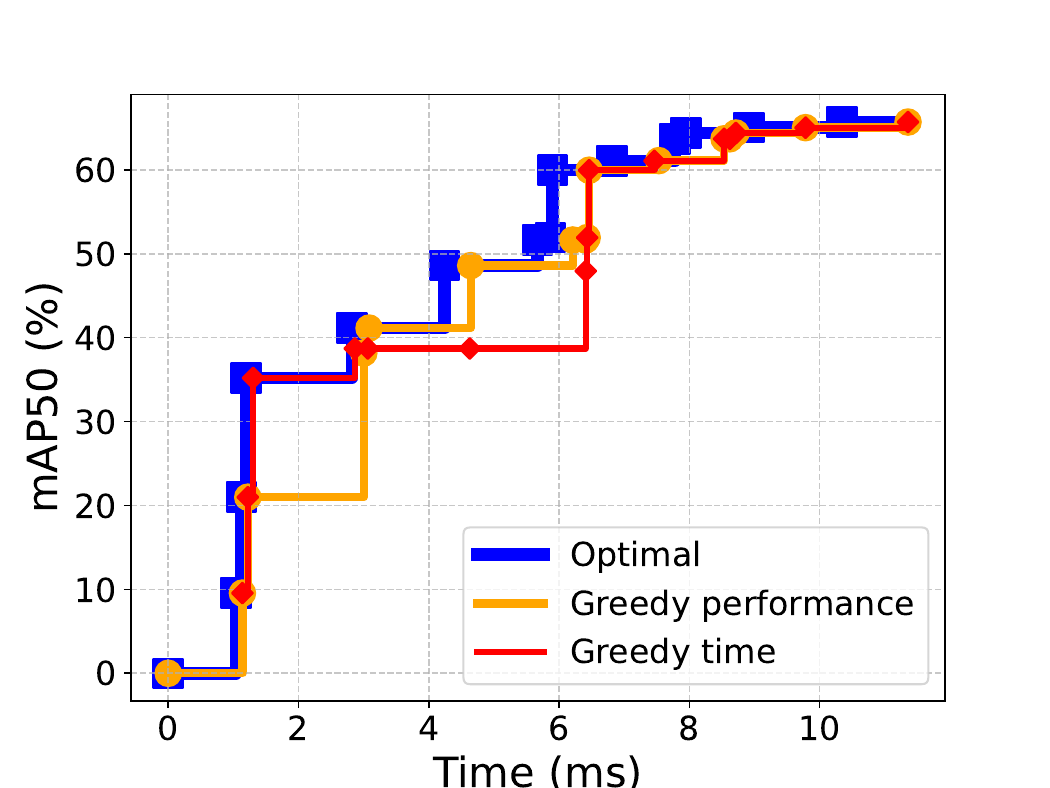}
    \caption{Optimal path selection performance compared to greedy heuristics, for \gelanmtransposed{}with 15 sub-exits, no pre-training.}
    \label{fig:pathselectionperformance}
\end{figure}

When comparing the three methods across \gelanmtransposed variants, exit choices mostly align. For nine sub-exits there is no difference. For 15 sub-exits, it increases to 5-6\% and 8-9\% without pre-training. Fig.~\ref{fig:pathselectionperformance} compares the methods for the no-pre-training variant, showing the greedy performance heuristic initially choosing an exit that is only slightly better but takes significantly longer. Similarly, greedy time chooses a marginally faster but significantly worse exit.

Using the Bellman-Ford algorithm, the optimal path selection terminates within seconds for our given model size once all the network's exits have been evaluated. Thus, for models with multiple paths, the optimal method offers significant performance gains over greedy heuristics at low pre-deployment cost, as shown in Figure~\ref{fig:pathselectionperformance}.

\section{Conclusion}
\label{sec:conclusion}
In this paper, we aim to introduce a structured exploration of anytime object detection networks, focusing on the popular YOLO architecture, introducing the AnytimeYOLO family of models. We give a theoretical measure of anytime quality to evaluate the performance of our proposed AnytimeYOLO models, enabling future work to compare anytime models. Our models are anytime capable at a moderate cost of the final accuracy. We highlight a trade-off between anytime quality and final exit quality, noting different factors. Our proposed \gelantransposed architecture achieves better anytime quality, especially at the beginning, sacrificing quality at the end.
We discuss how anytime models can be deployed, distinguishing between two types, soft and hard anytime inference. We propose methods to select the optimal order for a network with multiple execution orders and to select a subset of exits from an anytime model. We also present challenges of deploying anytime models, focusing on GPUs, emphasizing the lack of support for anytime inference in the current GPU frameworks. 

For future work, we plan to explore more features in addition to early-exits to achieve finer granularity. Furthermore, we will develop more efficient mechanisms that are compatible with existing inference frameworks, e.g., ONNXRuntime, TensorRT or Tensorflow Lite.

{
    \small
    \bibliographystyle{ieeenat_fullname}
    \bibliography{main}
}

\clearpage
\setcounter{page}{1}
\maketitlesupplementary

\appendix

\section{Implementation details}
\label{sec:suppl-implementation-details}
We shortly give implementation details of AnytimeYOLO, including the hyperparameters used for training, the architecture of both the medium variants \gelanm, \gelanmtransposed and tiny variants \gelant{}, \gelantransposed{}, and the list of sub-exits trained for each variant.

We use the hyperparameters of YOLOv9~\cite{DBLP:conf/eccv/WangYL24} for training all variants of AnytimeYOLO and list them in \cref{tab:hyperparameters}. A linear learning rate schedule is used with a warmup period of 3 epochs.

\begin{table}[h!]
    \centering
    \begin{tabular}{ll}
    \textbf{Parameter} & \textbf{Value} \\ \hline \hline
    epochs           & 500            \\
    optimizer        & SGD            \\
    initial learning rate               & 0.01           \\
    finish learning rate               & 0.0001           \\
    momentum          & 0.937          \\
    weight decay     & 0.0005         \\
    warmup epochs    & 3.0            \\
    warmup momentum  & 0.8            \\
    warmup bias lr  & 0.1            \\
    box loss gain               & 7.5            \\
    cls loss gain               & 0.5            \\
    obj loss gain              & 0.7            \\
    DFL loss gain               & 1.5            \\
    fl\_gamma         & 0.0            \\
    HSV$_H$ augmentation    & 0.015          \\
    HSV$_S$ augmentation    & 0.7            \\
    HSV$_V$ augmentation    & 0.4            \\
    translate augmentation & 0.1            \\
    scale augmentation     & 0.9            \\
    fliplr augmentation    & 0.5            \\
    mosaic augmentation    & 1.0            \\
    mixup augmentation     & 0.15           \\
    copy\_paste augmentation & 0.3          \\
    close mosaic epochs & 15             \\
    \end{tabular}
    \caption{Hyperparameters used for training of AnytimeYOLO.}
    \label{tab:hyperparameters}
\end{table}

The architecture of \gelant{} is shown in \cref{tab:backbone_head_summary}, the architecture of \gelantransposed{} is shown in \cref{tab:cfg_architecture}, the architecture of \gelanm{} is shown in \cref{tab:gelanm-architecture}, and the architecture of \gelanmtransposed{} is shown in \cref{tab:gelanmtransposed-architecture}. In the tables, the baseline configurations have the \emph{Detect} module as the only exit. We recall that an exit is composed of three sub-exits. In the case of \gelant{}, the exit is denoted as (15, 18, 21), corresponding to the sub-exits attached to the blocks with indices 15, 18 and 21, respectively. After the first three blocks, which serve as the initial backbone, these tables correspond to what is shown in \cref{fig:architecture-gelan-t} and \cref{fig:architecture-transposed}. For each row, the route column specifies the input blocks to the corresponding block. The filters, size and stride columns indicate the number of filters, kernel size and stride of the convolutions used. The CSP-ELAN and SPP-ELAN blocks are forms of GELAN block introduced in YOLOv9.

\begin{table*}[p]
    \centering
    \begin{tabular}{ccccccc}
        \textbf{Index} & \textbf{Module} & \textbf{Route} & \textbf{Filters} & \textbf{Depth} & \textbf{Size} & \textbf{Stride} \\ \hline
        0 & Conv & - & 16 & - & 3 & 2 \\ 
        1 & Conv & 0 & 32 & - & 3 & 2 \\ 
        2 & ELAN1 & 1 & 32, 32, 16 & - & - & - \\ 
        3 & Down & 2 & 64 & - & 3 & 2 \\ 
        4 & CSP-ELAN & 3 & 64, 64, 32 & 2,1 & - & 1 \\ 
        5 & Down & 4 & 96 & - & 3 & 2 \\ 
        6 & CSP-ELAN & 5 & 96, 96, 48 & 2,1 & - & 1 \\ 
        7 & Down & 6 & 128 & - & 3 & 2 \\ 
        8 & CSP-ELAN & 7 & 128, 128, 64 & 2,1 & - & 1 \\ 
        9 & SPP-ELAN & 8 & 128, 64 & 3,1 & - & - \\ 
        10 & Up & 9 & - & - & - & 2 \\ 
        11 & Concat & 10, 6 & - & - & - & - \\ 
        12 & CSP-ELAN & 11 & 96, 96, 48 & 2,1 & - & 1 \\ 
        13 & Up & 12 & - & - & - & 2 \\ 
        14 & Concat & 13, 4 & - & - & - & - \\ 
        15 & CSP-ELAN & 14 & 64, 64, 32 & 2,1 & - & 1 \\ 
        16 & Down & 15 & 48 & - & 3 & 2 \\ 
        17 & Concat & 16, 12 & - & - & - & - \\ 
        18 & CSP-ELAN & 17 & 96, 96, 48 & 2,1 & - & 1 \\ 
        19 & Down & 18 & 64 & - & 3 & 2 \\ 
        20 & Concat & 19, 9 & - & - & - & - \\ 
        21 & CSP-ELAN & 20 & 128, 128, 64 & 2,1 & - & 1 \\ 
        22 & Detect & 15, 18, 21 & - & - & - & - \\ 
    \end{tabular}
    \caption{Architecture of \gelant{} baseline.}
    \label{tab:backbone_head_summary}
\end{table*}
\begin{table*}[p]
    \centering
    \begin{tabular}{ccccccc}
        \textbf{Index} & \textbf{Module} & \textbf{Route} & \textbf{Filters} & \textbf{Depth} & \textbf{Size} & \textbf{Stride} \\ \hline
        0 & Conv & - & 16 & - & 3 & 2 \\ 
        1 & Conv & 0 & 32 & - & 3 & 2 \\ 
        2 & ELAN1 & 1 & 32, 32, 16 & - & - & - \\ 
        3 & Down & 2 & 64 & - & 3 & 2 \\ 
        4 & Down & 3 & 96 & - & 3 & 2 \\ 
        5 & Down & 4 & 128 & - & 3 & 2 \\ 
        6 & CSP-ELAN & 5 & 128, 128, 64 & 2,1 & - & 1 \\ 
        7 & CSP-ELAN & 4 & 96, 96, 48 & 2,1 & - & 1 \\ 
        8 & CSP-ELAN & 3 & 64, 64, 32 & 2,1 & - & 1 \\ 
        9 & SPP-ELAN & 6 & 128, 64 & 3,1 & - & - \\ 
        10 & Up & 9 & - & - & - & 2 \\ 
        11 & Concat & 10, 7 & - & - & - & - \\ 
        12 & Up & 11 & - & - & - & 2 \\ 
        13 & Concat & 12, 8 & - & - & - & - \\ 
        14 & CSP-ELAN & 11 & 96, 96, 48 & 2,1 & - & 1 \\ 
        15 & CSP-ELAN & 13 & 64, 64, 32 & 2,1 & - & 1 \\ 
        16 & Down & 15 & 48 & - & 3 & 2 \\ 
        17 & Concat & 16, 14 & - & - & - & - \\ 
        18 & Down & 17 & 64 & - & 3 & 2 \\ 
        19 & Concat & 18, 9 & - & - & - & - \\ 
        20 & CSP-ELAN & 19 & 128, 128, 64 & 2,1 & - & 1 \\ 
        21 & CSP-ELAN & 17 & 96, 96, 48 & 2,1 & - & 1 \\ 
        22 & Detect & 15, 21, 20 & - & - & - & - \\ 
    \end{tabular}
    \caption{Architecture of \gelantransposed{} baseline.}
    \label{tab:cfg_architecture}
\end{table*}

\begin{table*}[p]
    \centering
    \begin{tabular}{ccccccc}
    \textbf{Index} & \textbf{Module} & \textbf{Route} & \textbf{Filters} & \textbf{Depth} & \textbf{Size} & \textbf{Stride} \\ \hline
    0 & Conv & - & 32 & - & 3 & 2 \\
    1 & Conv & 0 & 64 & - & 3 & 2 \\
    2 & CSP-ELAN & 1 & 128, 128, 64 & 2,1 & - & 1 \\
    3 & Down & 2 & 240 & - & 3 & 2 \\
    4 & CSP-ELAN & 3 & 240, 240, 120 & 2,1 & - & 1 \\
    5 & Down & 4 & 360 & - & 3 & 2 \\
    6 & CSP-ELAN & 5 & 360, 360, 180 & 2,1 & - & 1 \\
    7 & Down & 6 & 480 & - & 3 & 2 \\
    8 & CSP-ELAN & 7 & 480, 480, 240 & 2,1 & - & 1 \\
    9 & SPP-ELAN & 8 & 480, 240 & 3,1 & - & - \\
    10 & Up & 9 & - & - & - & 2 \\
    11 & Concat & 10, 6 & - & - & - & - \\
    12 & CSP-ELAN & 11 & 360, 360, 180 & 2,1 & - & 1 \\
    13 & Up & 12 & - & - & - & 2 \\
    14 & Concat & 13, 4 & - & - & - & - \\
    15 & CSP-ELAN & 14 & 240, 240, 120 & 2,1 & - & 1 \\
    16 & Down & 15 & 180 & - & 3 & 2 \\
    17 & Concat & 16, 12 & - & - & - & - \\
    18 & CSP-ELAN & 17 & 360, 360, 180 & 2,1 & - & 1 \\
    19 & Down & 18 & 240 & - & 3 & 2 \\
    20 & Concat & 19, 9 & - & - & - & - \\
    21 & CSP-ELAN & 20 & 480, 480, 240 & 2,1 & - & 1 \\
    22 & Detect & 15, 18, 21 & - & - & - & - \\
    \end{tabular}
    \caption{Architecture of \gelanm{} baseline.}
    \label{tab:gelanm-architecture}
\end{table*}

\begin{table*}[p]
    \centering
    \begin{tabular}{ccccccc}
    \textbf{Index} & \textbf{Module} & \textbf{Route} & \textbf{Filters} & \textbf{Depth} & \textbf{Size} & \textbf{Stride} \\ \hline
    0 & Conv & - & 32 & - & 3 & 2 \\
    1 & Conv & 0 & 64 & - & 3 & 2 \\
    2 & CSP-ELAN & 1 & 128, 128, 64 & 2,1 & - & 1 \\
    3 & Down & 2 & 240 & - & 3 & 2 \\
    4 & Down & 3 & 360 & - & 3 & 2 \\
    5 & Down & 4 & 480 & - & 3 & 2 \\
    6 & CSP-ELAN & 5 & 480, 480, 240 & 2,1 & - & 1 \\
    7 & CSP-ELAN & 3 & 360, 360, 180 & 2,1 & - & 1 \\
    8 & CSP-ELAN & 1 & 240, 240, 120 & 2,1 & - & 1 \\
    9 & SPPELAN & 6 & 480, 240 & 3,1 & - & - \\    
    10 & Up & 9 & - & - & - & 2 \\
    11 & Concat & 10, 7 & - & - & - & - \\
    12 & Up & 11 & - & - & - & 2 \\
    13 & Concat & 12, 8 & - & - & - & - \\
    14 & CSP-ELAN & 11 & 360, 360, 180 & 2,1 & - & 1 \\
    15 & CSP-ELAN & 13 & 240, 240, 120 & 2,1 & - & 1 \\
    16 & Down & 15 & 180 & - & 3 & 2 \\
    17 & Concat & 16, 14 & - & - & - & - \\
    18 & Down & 17 & 240 & - & 3 & 2 \\
    19 & Concat & 18, 9 & - & - & - & - \\
    20 & CSP-ELAN & 19 & 480, 480, 240 & 2,1 & - & 1 \\
    21 & CSP-ELAN & 17 & 360, 360, 180 & 2,1 & - & 1 \\
    22 & Detect & 15, 21, 20 & - & - & - & - \\ 
\end{tabular}
    \caption{Architecture of \gelanmtransposed{} baseline.}
    \label{tab:gelanmtransposed-architecture}
\end{table*}

The full list of trained sub-exits for \gelant{} is [(15, 18, 21),
    (15, 17, 20),
    (15, 12, 9),
    (14, 11, 9),
    (4, 6, 8),
    (3, 5, 7)]. The list of sub-exits for our \gelant{} variant with only 9 sub-exits is [(15, 18, 21),
    (14, 11, 9),
    (4, 5, 7)]

As mentioned in~\cref{sec:training}, for training, the loss is computed for a set of sub-exits with one sub-exit of each resolution. The sub-exits in each tuple are those that are trained together which is why some are redundant. However, during inference different combinations can be used. As an example, the exit (4,6,7) can be used, even if only (3,5,7) and (4,6,8) were specifically trained for.

The full list of trained sub-exits for \gelantransposed{} is [(15, 21, 20),
(15, 17, 19),
(15, 14, 9),
(13, 11, 9),
(8, 7, 6),
(3, 4, 5)]. The list of sub-exits for our \gelantransposed{} variant with only 9 sub-exits is [(15, 21, 20),
(13, 11, 9),
(3, 4, 5)].

\section{Additional evaluation results}
\label{sec:suppl-full-evaluation}
We present additional results of our evaluation of AnytimeYOLO variants, both for the tiny and medium variants. We show performance plots for all variants, an additional performance table with the AP$_{50:90}$ quality, the time each layer takes to execute and a comparison with the naive approach at different batch sizes.

Table.~\ref{tab:tiny-performance} shows the performance comparison of the tiny variant, \gelant{} and \gelantransposed{}. It is structured in the same fashion as the medium variant Table~\ref{tab:performance}, showing the performance of the variants with and without pre-training, with 9 and 15 sub-exits. The trends that can be observed are similar to the medium variant, though the loss of performance by adding exits is less pronounced.

\begin{table}
    \centering
    \scalebox{0.7}{
    \begin{tabular}{l|p{1cm}|p{0.7cm}|p{0.9cm}|p{1.2cm}|p{1cm}|p{1.1cm}|p{1.6cm}}
    Name & Pretrain & Exits & $Q_{{AP}_{50}}$ & 
    $Q_{AP_{50}SE}$ & 
    AP$_{50}$ & 
    $\max \Delta$ & 
    MSize \\ \hline \hline
    YOLO9-t & - & - & - & - & 53.1 & 9.92 ms & 9.2 MB \\

    \gelant & - & - & - & - & 51.34 & 9.98 ms & 9.2 MB \\

    \gelantransposed & - & - & - & - & 49.4 & 9.90 ms & 10.52 MB \\

    \hline
    
    \gelant & 0 & 9 & 33.7 & 6.01 & 49.1 & 2.84 ms & 19.9 MB\\
    \gelant & 0 & 15 & 34.79 & 5.65 & 49.84 & \textbf{2.46 ms} & 32.07 MB \\
    \gelant & 200 & 9 & 33.76 & 6.1 & 50.29 & 2.84 ms & 19.9 MB \\
    \gelant & 200 & 15 & 34.6 & 5.81 & \textbf{50.54} & \textbf{2.46 ms} & 32.07 MB \\
    \gelantransposed & 0 & 9 & 34.92 & 4.02 & 46.91 & 2.78 ms & 19.29 MB \\
    \gelantransposed & 0 & 15 & \textbf{36.14} & \textbf{3.7} & 46.98 & 2.54 ms & 34.56 MB \\
    \gelantransposed & 200 & 9 & 35.09 & 4.1 & 48.22 & 2.78 ms & 19.29 MB \\
    \gelantransposed & 200 & 15 & 35.51 & 3.98 & 47.62 & 2.54 ms & 34.56 MB \\
    \end{tabular}}
    \caption{Performance Comparison of \gelant{} and \gelantransposed{}}
    \label{tab:tiny-performance}
\end{table}

\begin{figure*}
    \begin{subfigure}[b]{0.5\textwidth}
    \centering
    \includegraphics[width=0.90\columnwidth]{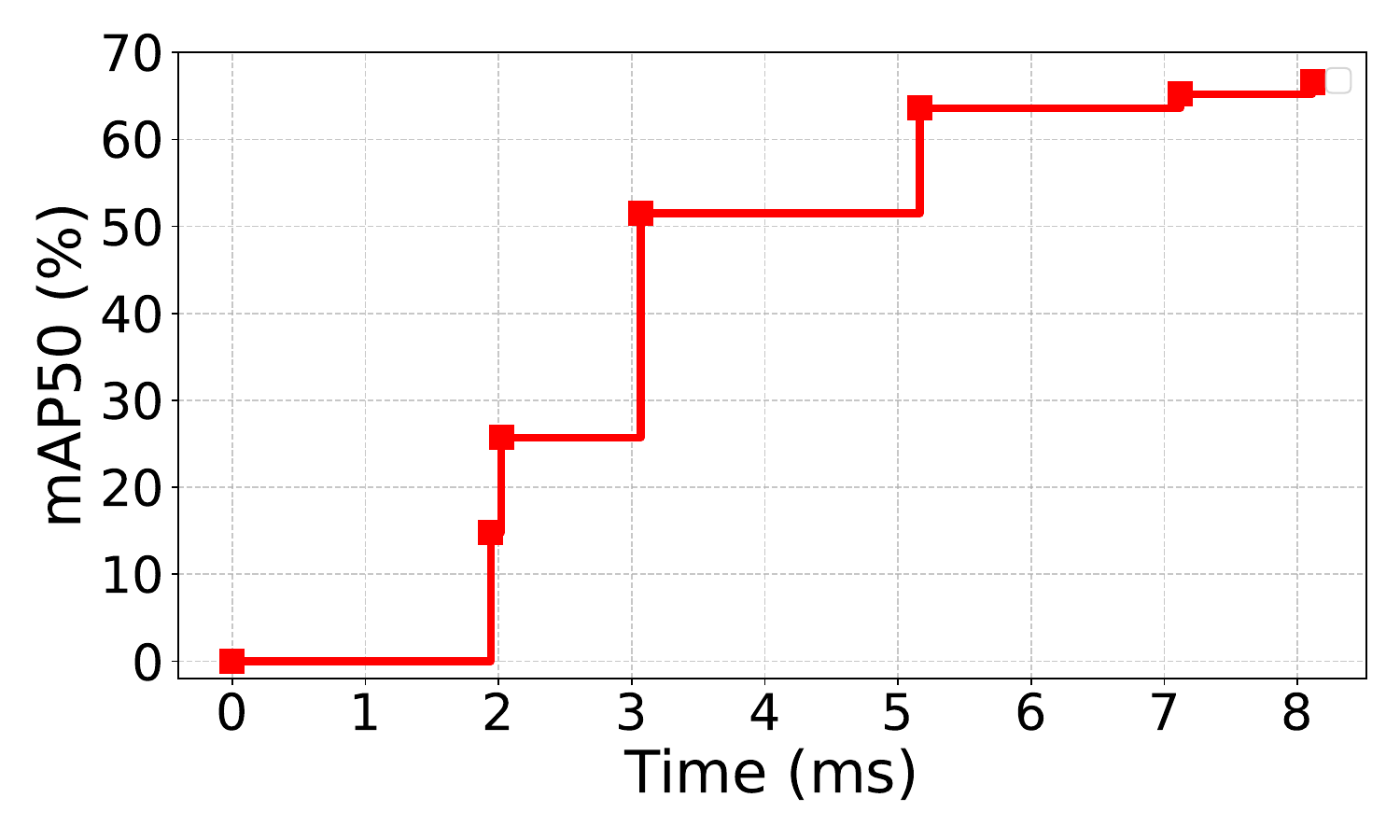}
    \caption{\gelanm{} with 9 exits.}
    \label{fig:gelan-m-three-exits}
    \end{subfigure}
    \begin{subfigure}[b]{0.5\textwidth}
    \centering
    \includegraphics[width=0.90\columnwidth]{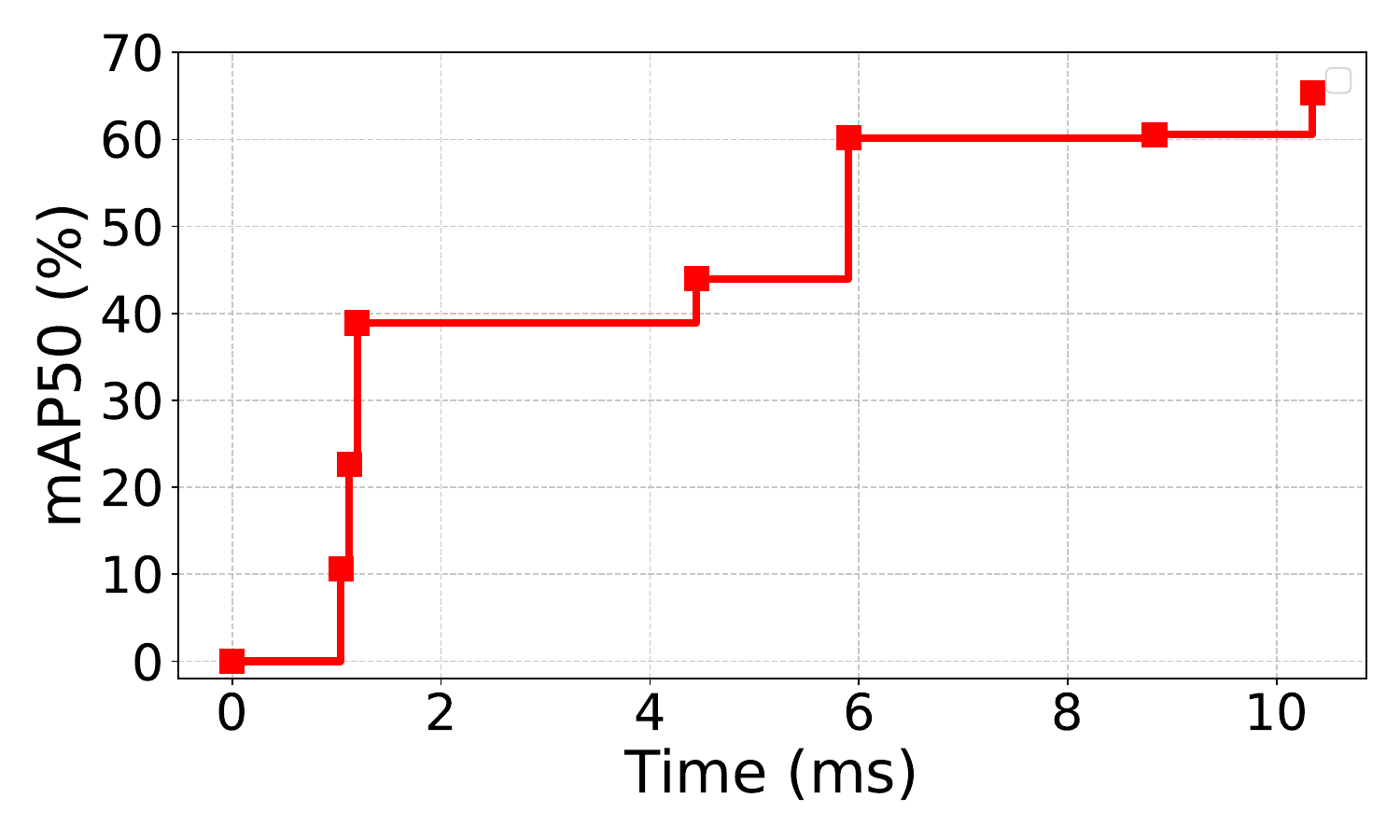}
    \caption{\gelanmtransposed{} with 9 exits.}
    \label{fig:gelan-m-transposed-three-exits}
    \end{subfigure}
    \begin{subfigure}[b]{0.5\textwidth}
    \centering
    \includegraphics[width=0.90\columnwidth]{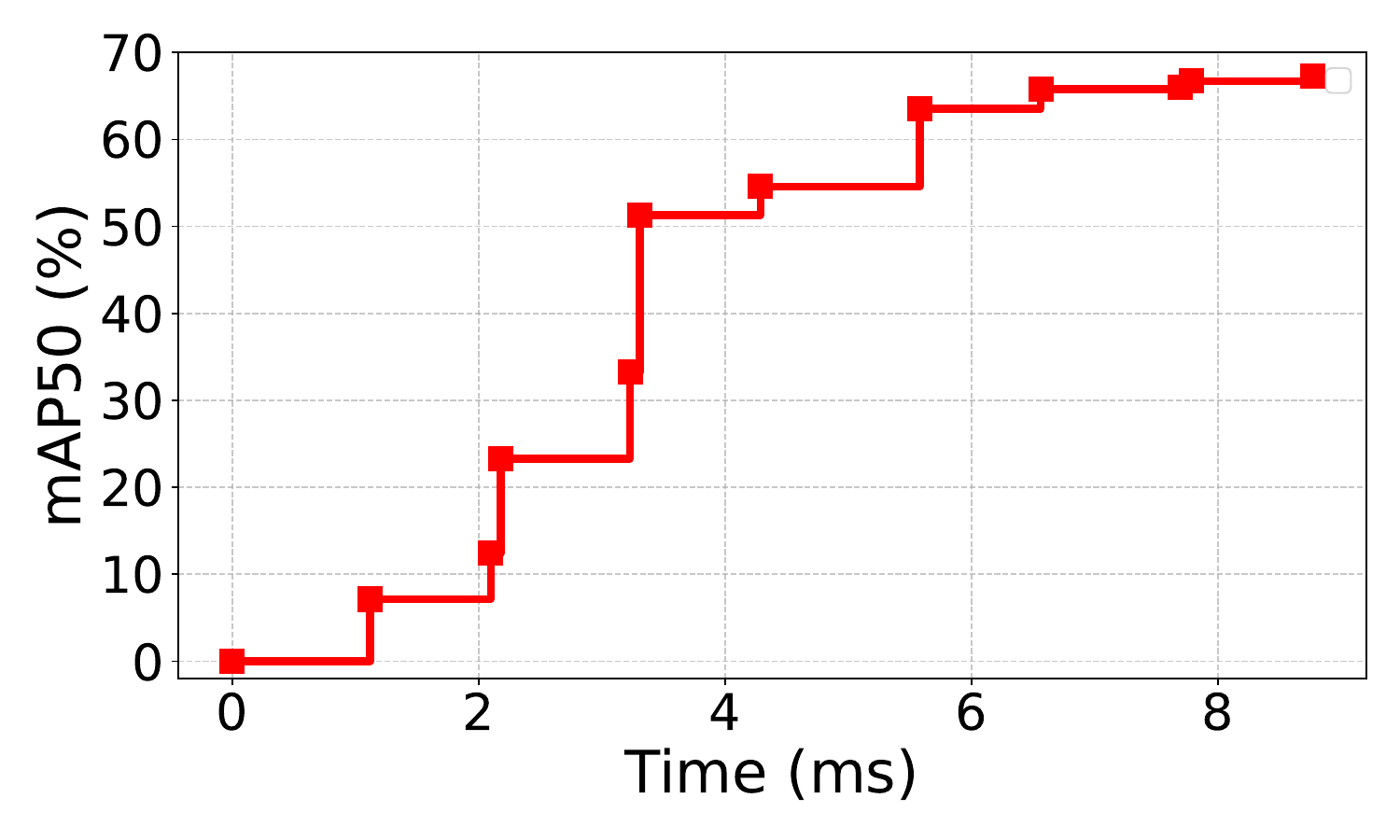}
    \caption{\gelanm{} with 15 exits.}
    \label{fig:gelan-m-six-exits}
    \end{subfigure}
    \begin{subfigure}[b]{0.5\textwidth}
    \centering
    \includegraphics[width=0.90\columnwidth]{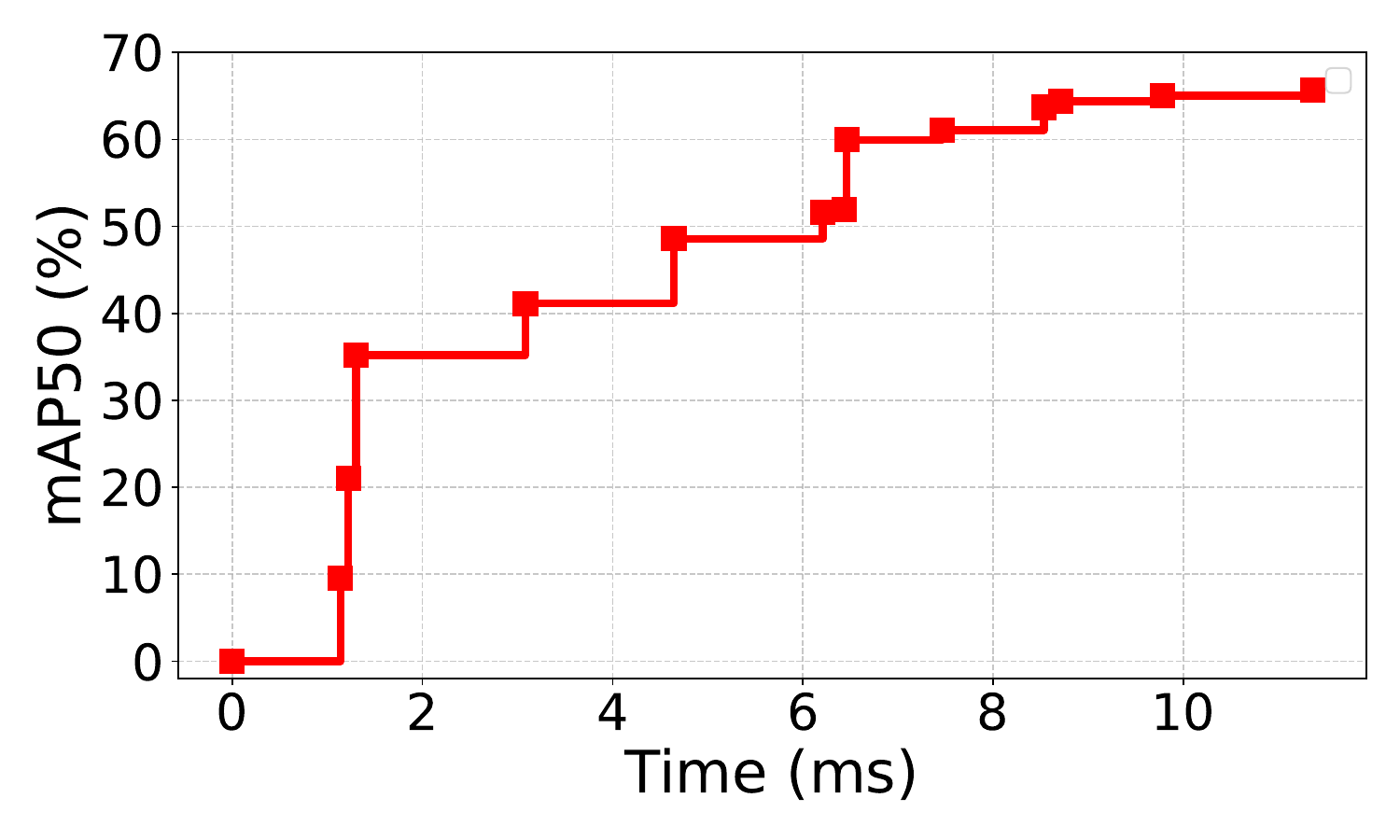}
    \caption{\gelanmtransposed{} with 15 exits.}
    \label{fig:gelan-m-transposed-six-exits}
    \end{subfigure}
    \begin{subfigure}[b]{0.5\textwidth}
    \centering
    \includegraphics[width=0.90\columnwidth]{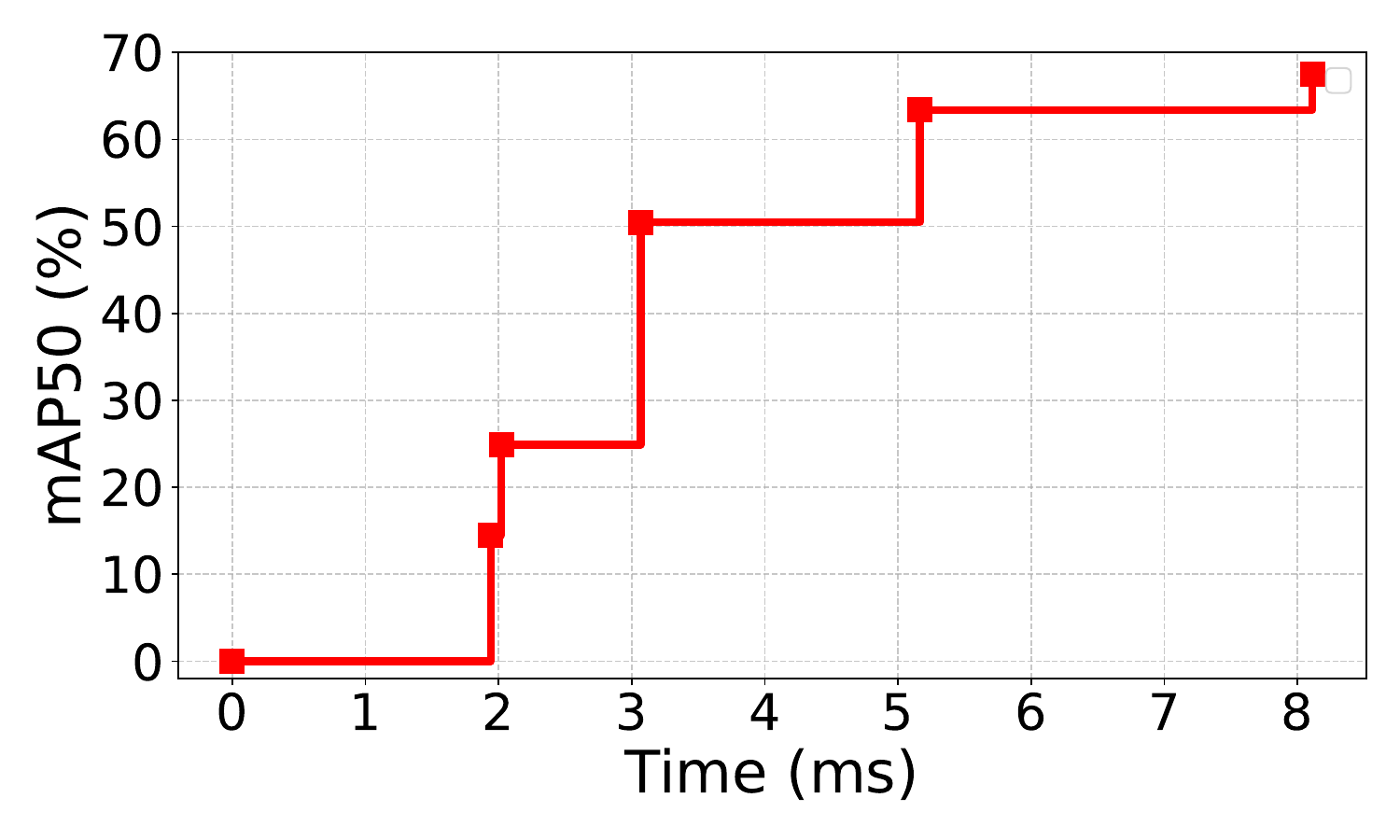}
    \caption{\gelanm{} with 9 exits and 200 epochs pre-training.}
    \label{fig:gelan-m-three-exits-200}
    \end{subfigure}
    \begin{subfigure}[b]{0.5\textwidth}
    \centering
    \includegraphics[width=0.90\columnwidth]{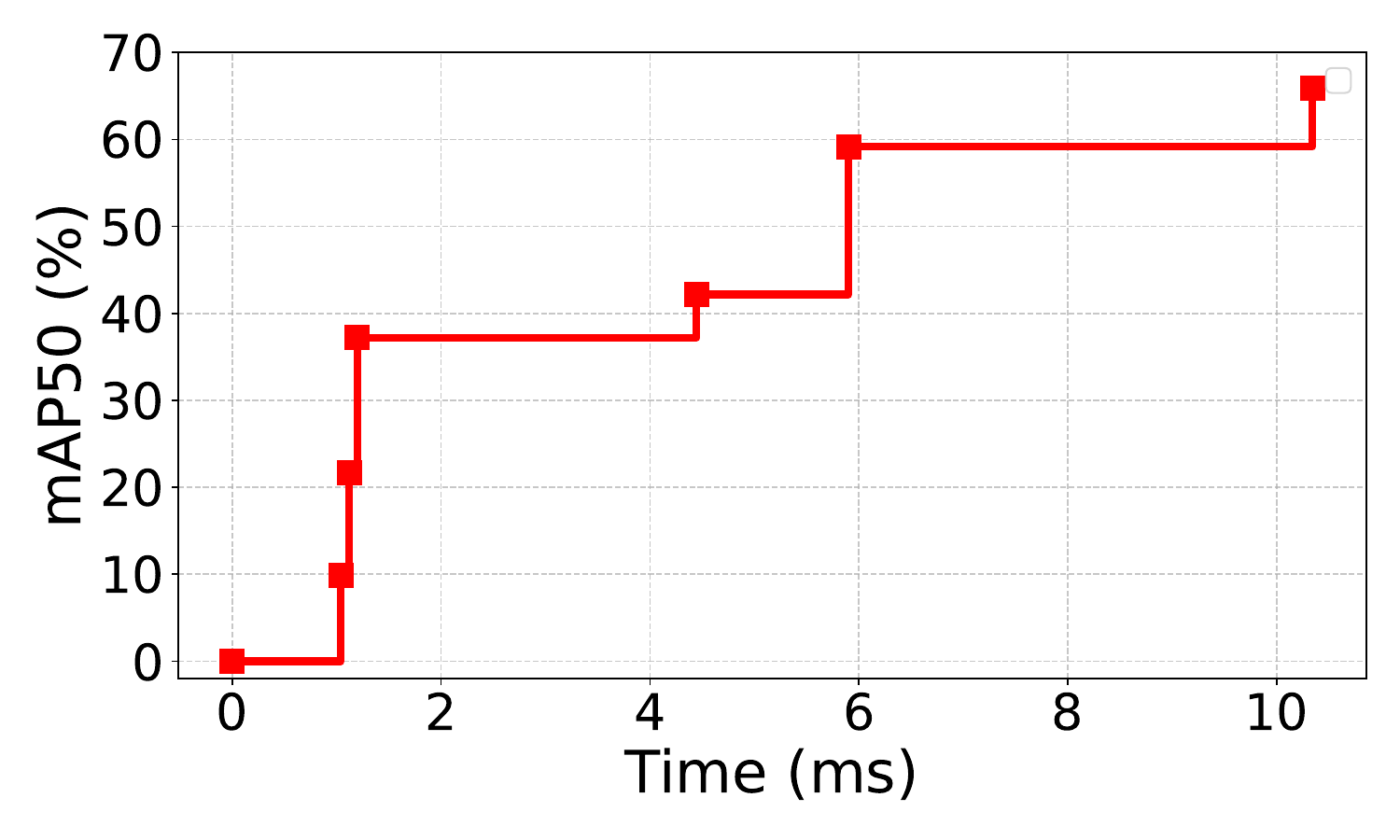}
    \caption{\gelanmtransposed{} with 9 exits and 200 epochs pre-training.}
    \label{fig:gelan-m-transposed-three-exits-200}
    \end{subfigure}
    \begin{subfigure}[b]{0.5\textwidth}
    \centering
    \includegraphics[width=0.90\columnwidth]{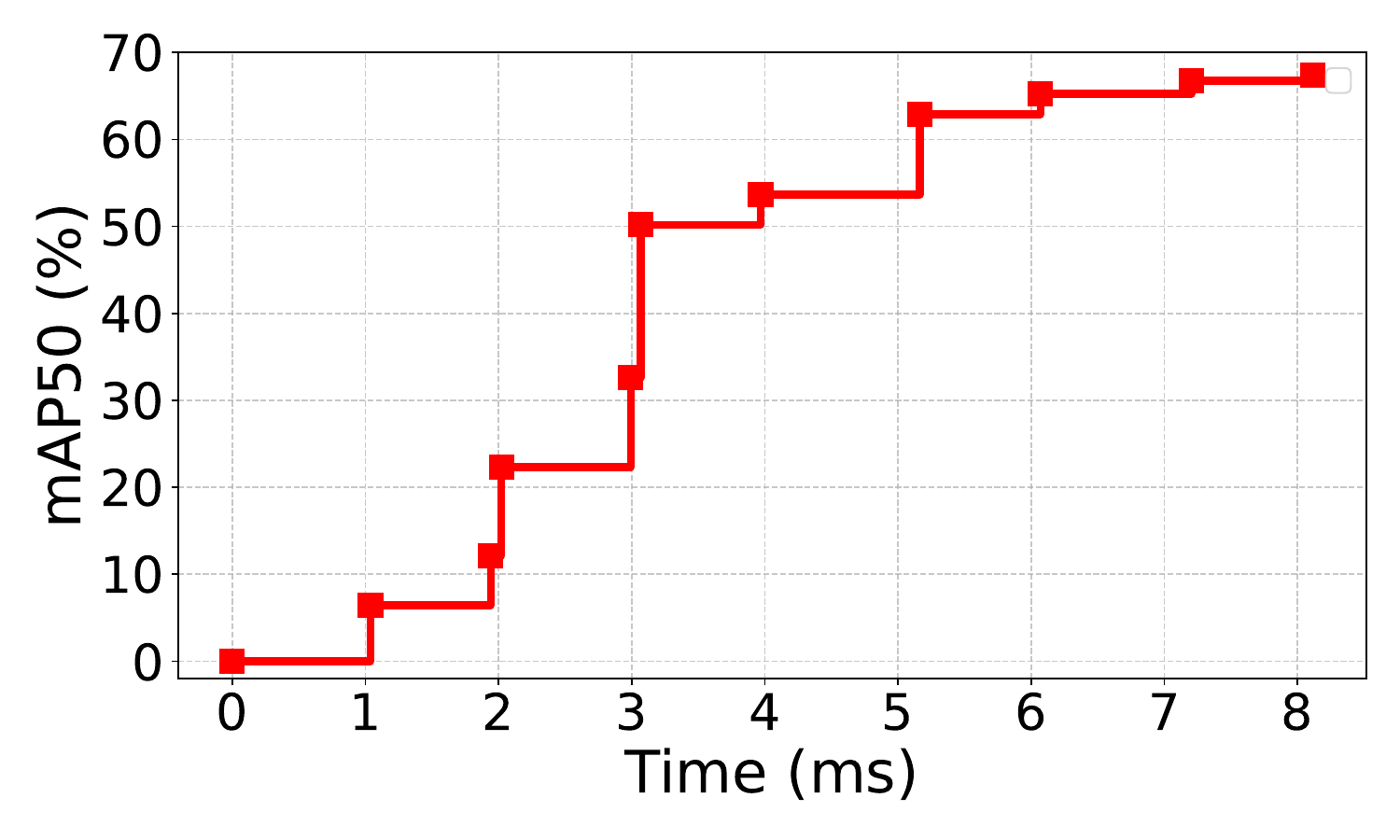}
    \caption{\gelanm{} with 15 exits and 200 epochs pre-training.}
    \label{fig:gelan-m-six-exits-200}
    \end{subfigure}
    \begin{subfigure}[b]{0.5\textwidth}
    \centering
    \includegraphics[width=0.90\columnwidth]{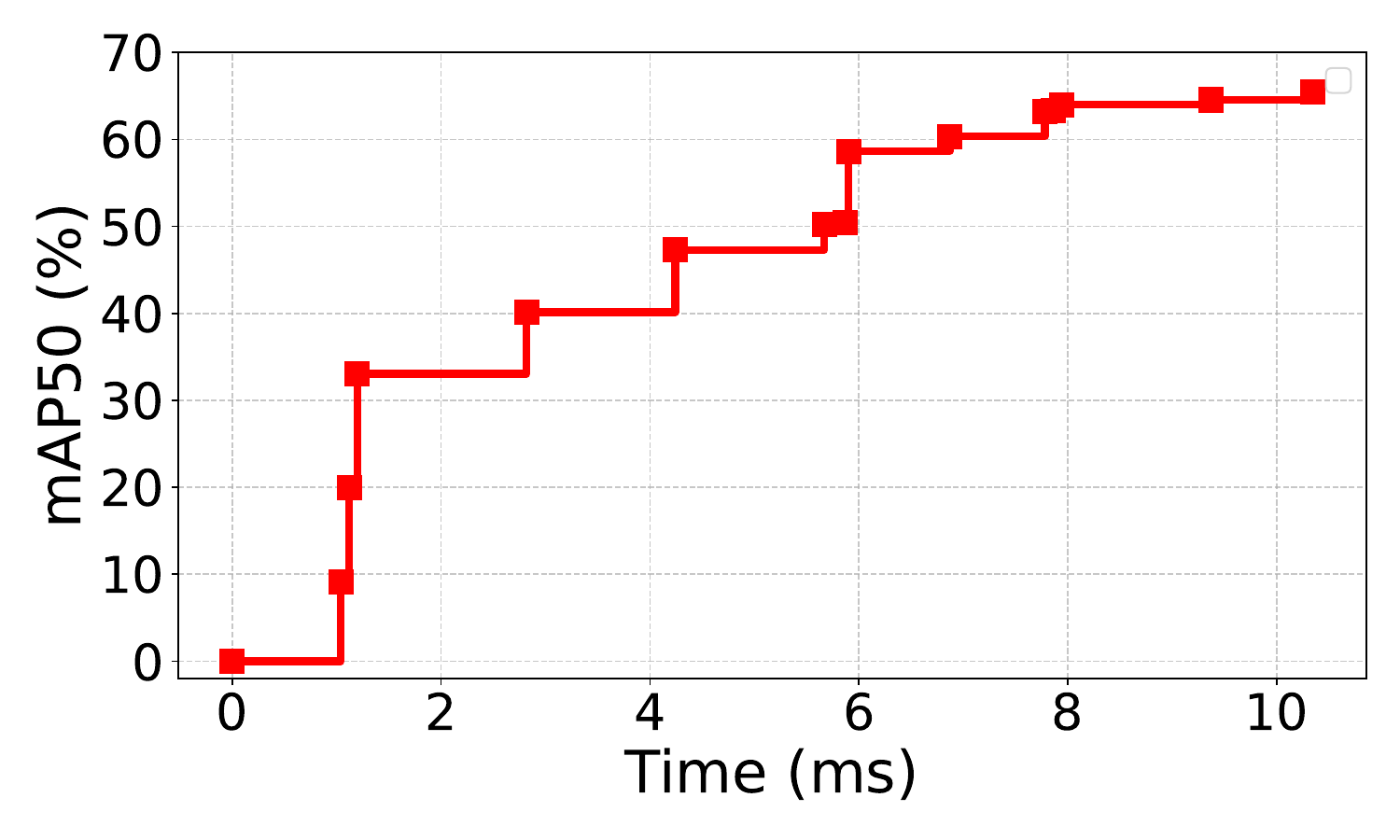}
    \caption{\gelanmtransposed{} with 15 exits and 200 epochs pre-training.}
    \label{fig:gelan-m-transposed-six-exits-200}
    \end{subfigure}
    \caption{Performance of AnytimeYOLO \gelanm{} and \gelanmtransposed{} variants.}
    \label{fig:full-evaluation-m}
 \end{figure*}

\begin{figure*}
    \begin{subfigure}[b]{0.5\textwidth}        
        \centering
        \includegraphics[width=0.90\columnwidth]{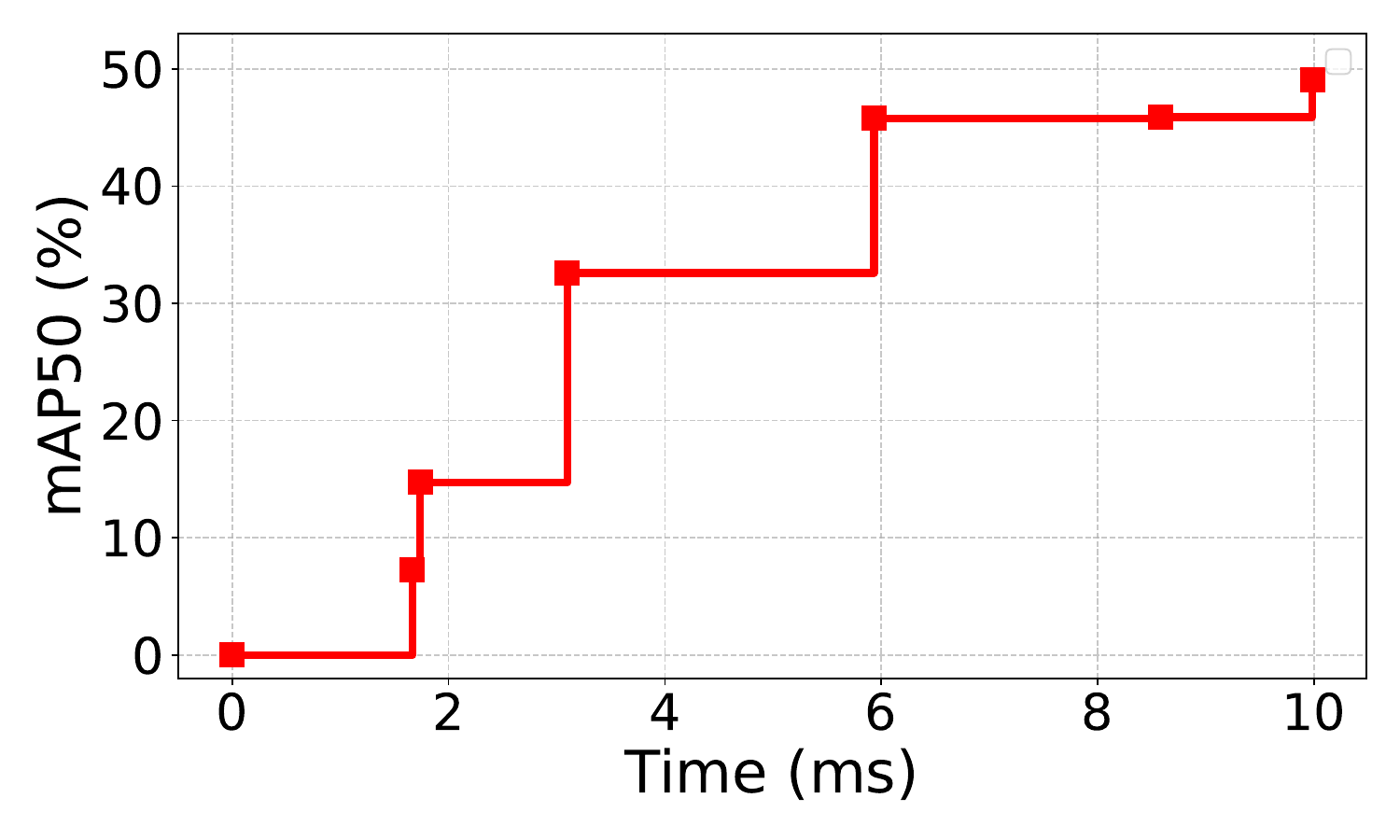}
        \caption{\gelant{} with 9 exits.}
        \label{fig:gelan-t-three-exits}
    \end{subfigure}
    \begin{subfigure}[b]{0.5\textwidth}
        \centering
        \includegraphics[width=0.90\columnwidth]{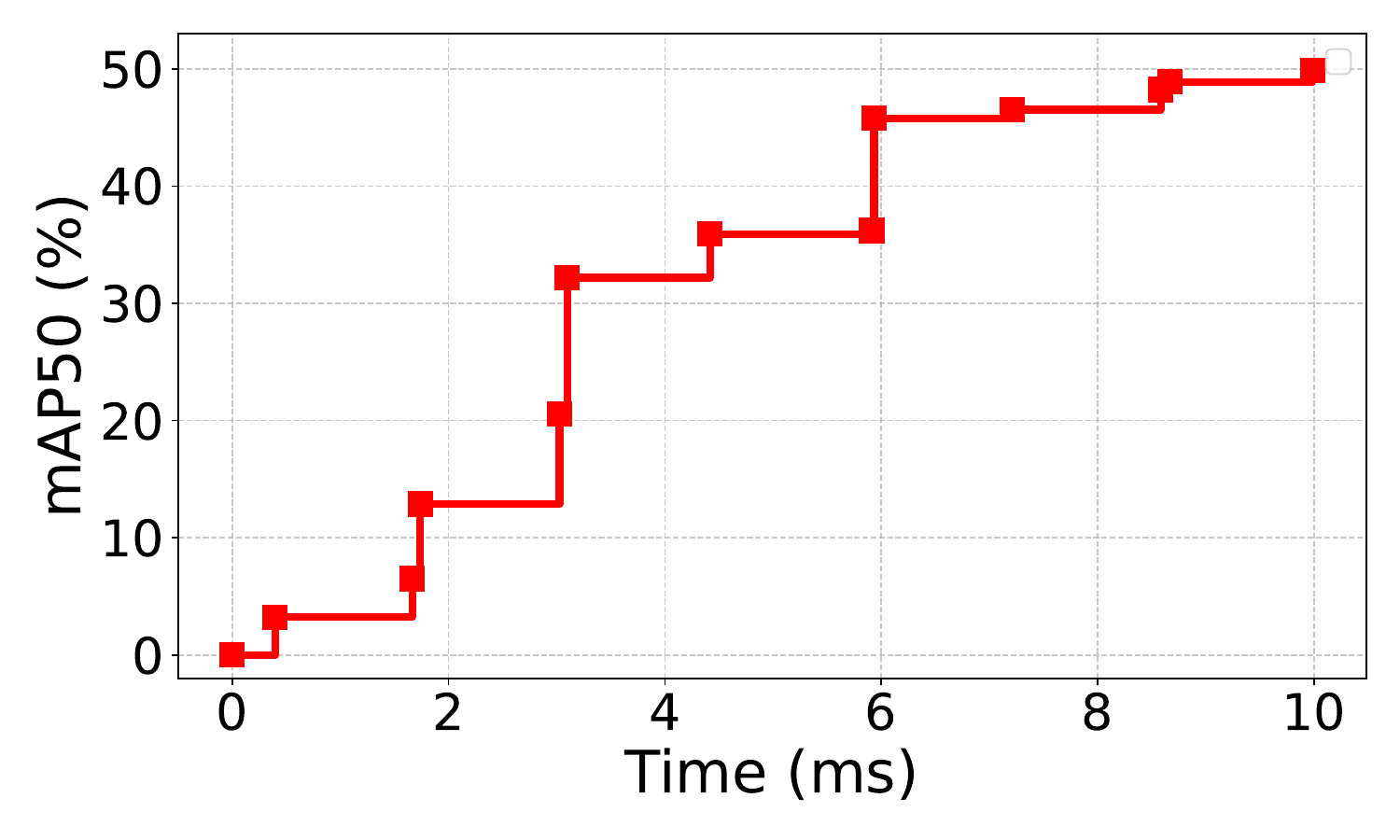}
        \caption{\gelant{} with 15 exits.}
        \label{fig:gelan-t-six-exits}
    \end{subfigure}
    \begin{subfigure}[b]{0.5\textwidth}
        \centering
        \includegraphics[width=0.90\columnwidth]{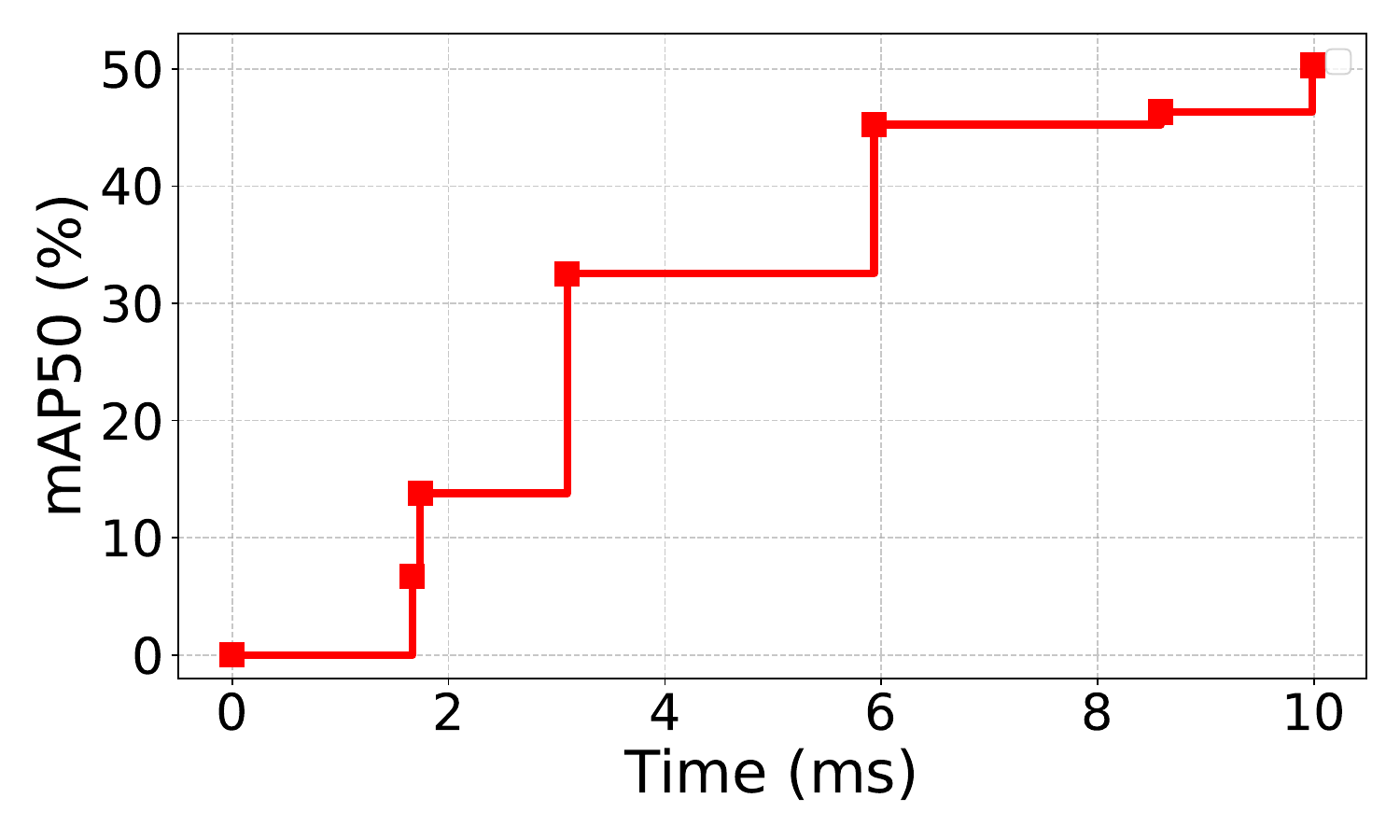}
        \caption{\gelant{} with 9 exits and 200 epochs pre-training.}
        \label{fig:gelan-t-three-exits-200}
    \end{subfigure}
    \begin{subfigure}[b]{0.5\textwidth}
        \centering
        \includegraphics[width=0.90\columnwidth]{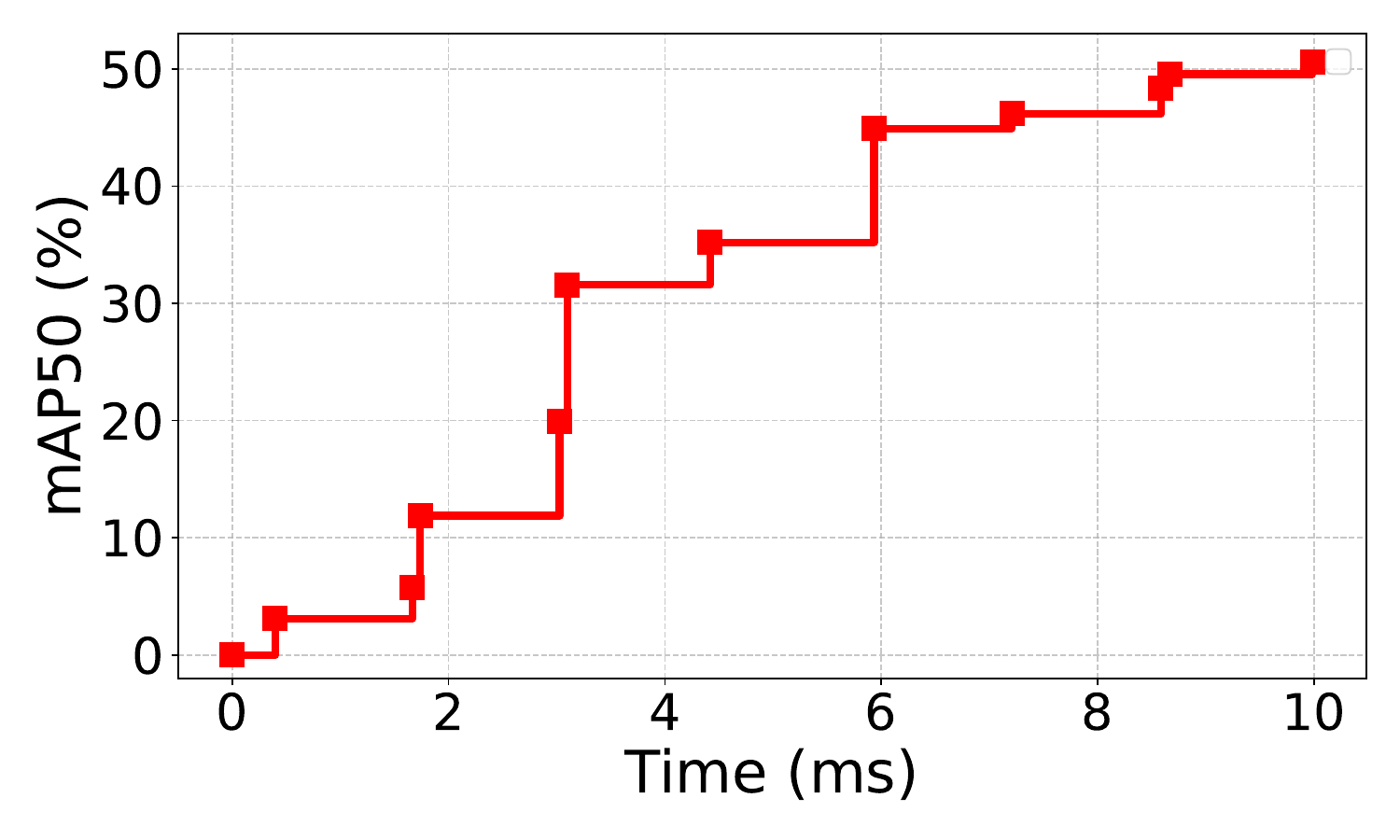}
        \caption{\gelant{} with 15 exits and 200 epochs pre-training.}
        \label{fig:gelan-t-six-exits-200}
    \end{subfigure}
    \begin{subfigure}[b]{0.5\textwidth}
        \centering
        \includegraphics[width=0.90\columnwidth]{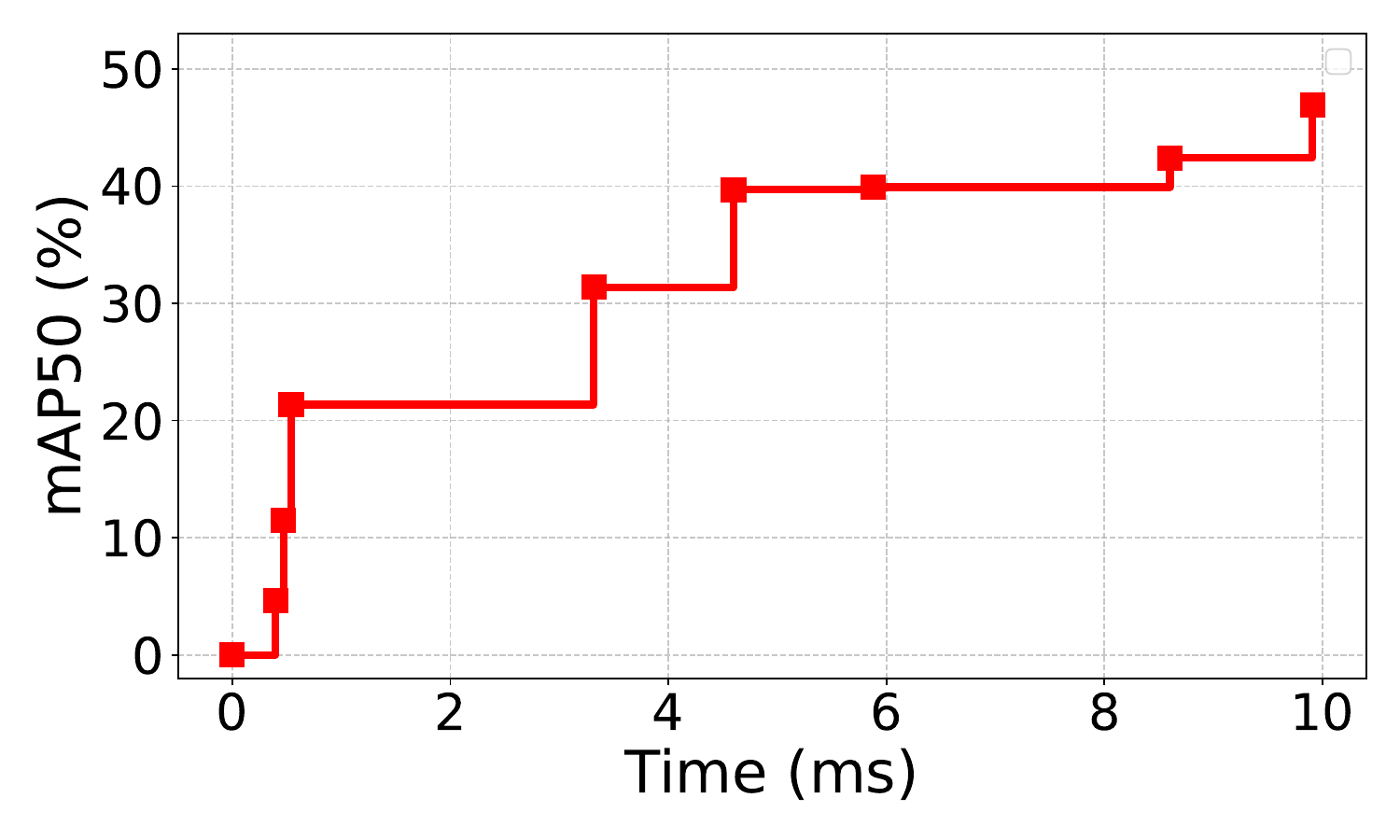}
        \caption{\gelantransposed{} with 9 exits.}
        \label{fig:transposed-three-exits}
    \end{subfigure}
    \begin{subfigure}[b]{0.5\textwidth}
        \centering
        \includegraphics[width=0.90\columnwidth]{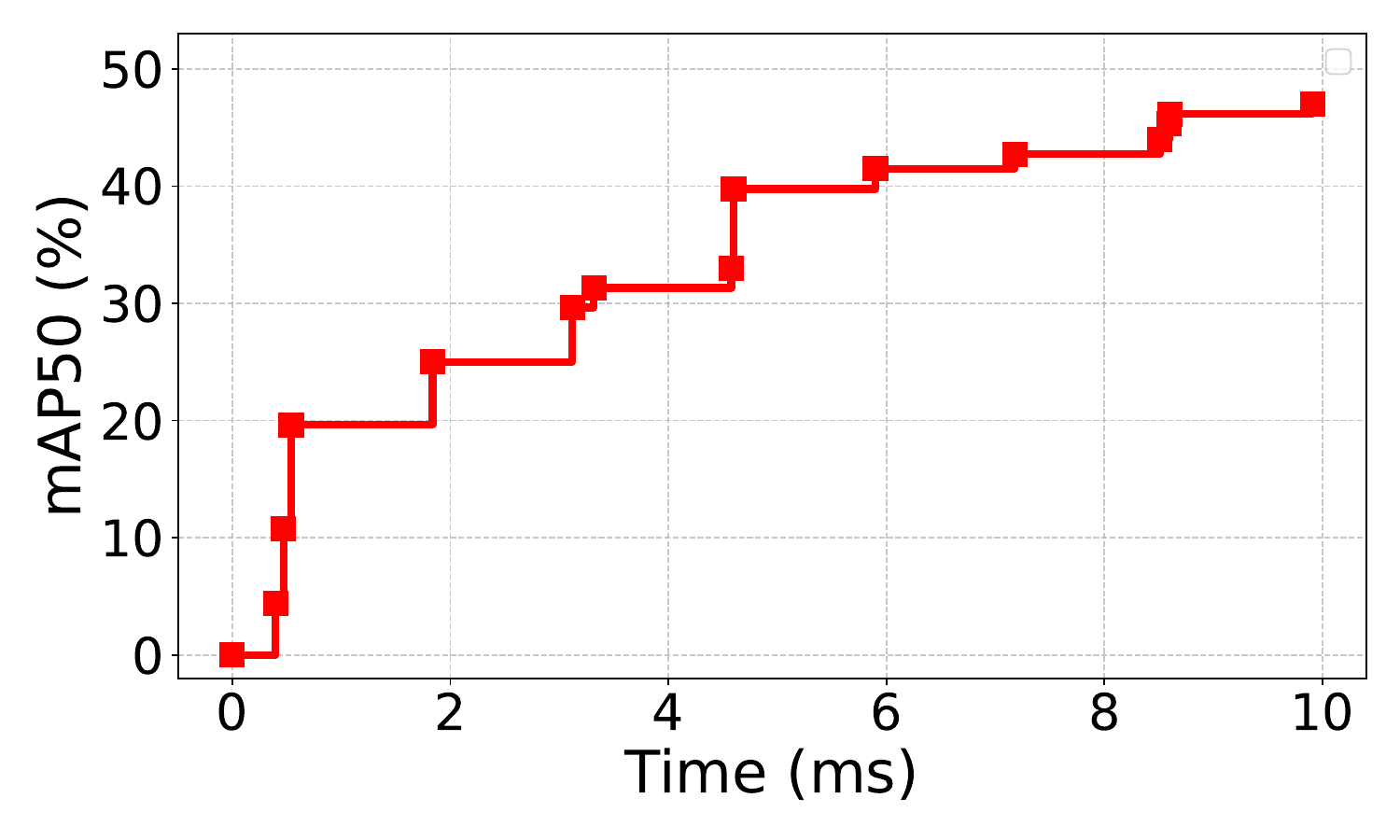}
        \caption{\gelantransposed{} with 15 exits.}
        \label{fig:transposed-six-exits}
    \end{subfigure}
    \begin{subfigure}[b]{0.5\textwidth}
        \centering
        \includegraphics[width=0.90\columnwidth]{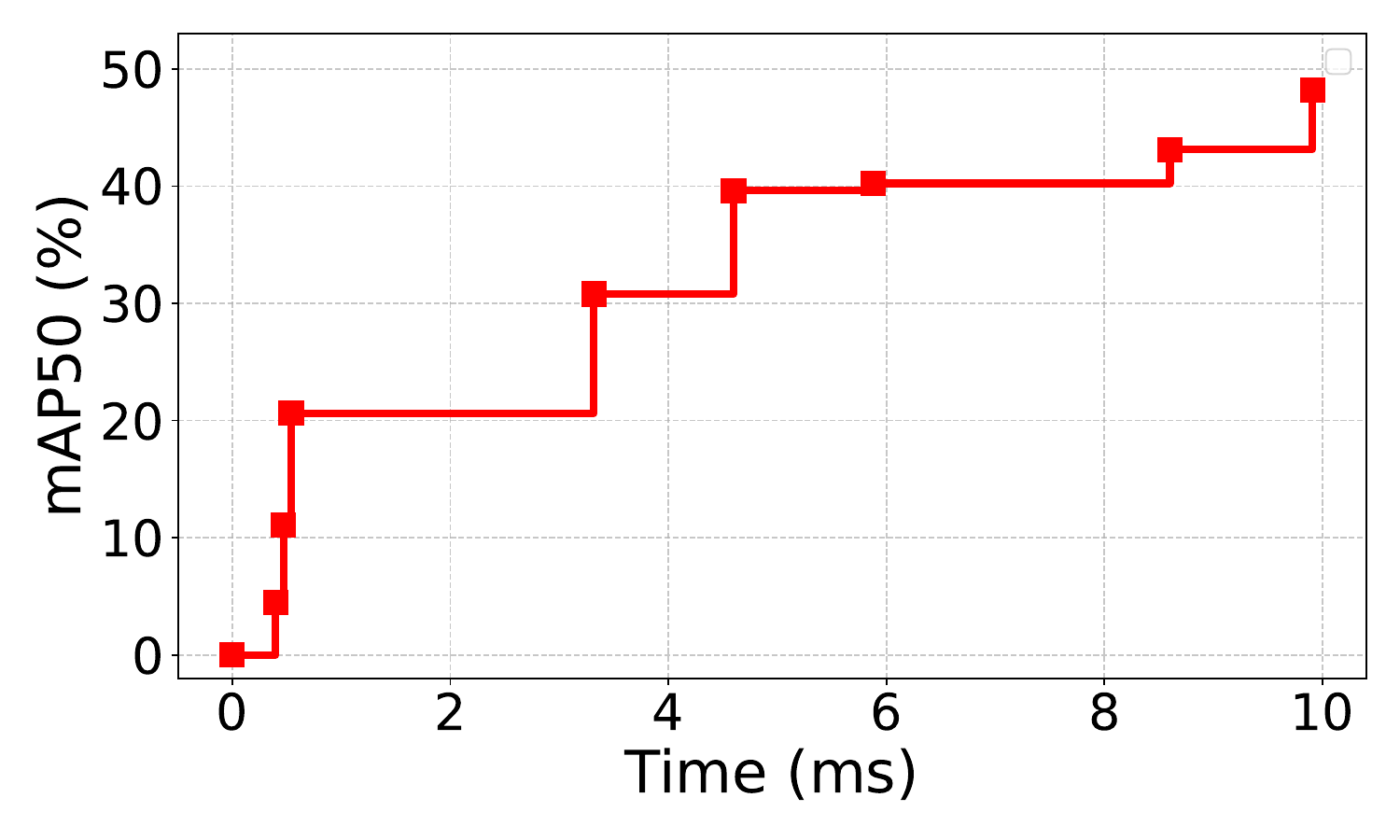}
        \caption{\gelantransposed{} with 9 exits and 200 epochs pre-training.}
        \label{fig:transposed-three-exits-200}
    \end{subfigure}
    \begin{subfigure}[b]{0.5\textwidth}
        \centering
        \includegraphics[width=0.90\columnwidth]{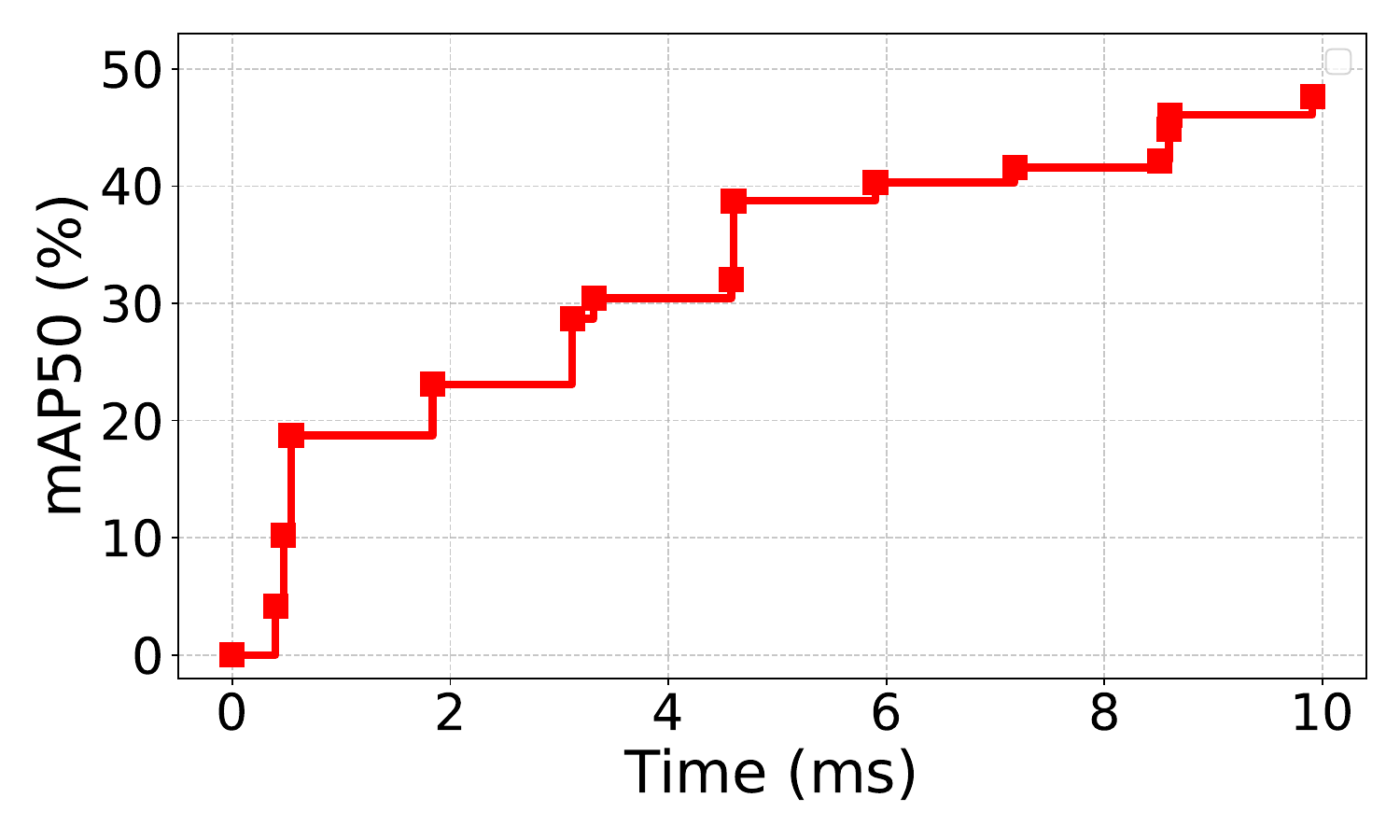}
        \caption{\gelantransposed{} with 15 exits and 200 epochs pre-training.}
        \label{fig:transposed-six-exits-200}
    \end{subfigure}
    \caption{Performance of AnytimeYOLO \gelant{} and \gelantransposed{} variants.}
    \label{fig:full-evaluation}
\end{figure*}

\cref{fig:full-evaluation-m} and \cref{fig:full-evaluation} show the performance plots of our medium and tiny AnytimeYOLO variants. The plots show the performance of the AnytimeYOLO variants for \gelant{}, \gelantransposed{}, \gelanm and \gelanmtransposed with 9 and 15 sub-exits, with and without 200 epochs of pre-training. We note that, e.g., 9 sub-exits do not correspond to 9 exit points as in some cases, to ensure monotonicity, some possible combinations are skipped. As an example, while a combination between the exits (4, 5, 7) and (14, 11, 9) such as (4, 5, 9) is possible, it was skipped due to having worse performance than (4, 5, 7).  
The plots show the AP$_{50}$ quality of the variants as a function of the inference time. One can see the higher granularity of the variants with 15 exits compared to the variants with 9 exits which results in a much better anytime quality. A further observation is that the variants with 9 exits have a slightly better performance at the beginning of the inference, while the variants with 15 exits have a slightly better performance at the end of the inference. The same can be seen for the variants with 200 epochs of pre-training, which have slightly better performance at the end but slightly worse performance at the beginning.

In \cref{tab:gelant-latency}, \cref{tab:gelantransposed-latency},
\cref{tab:gelanm-latency}, and \cref{tab:gelanmtransposed-latency} we show the average time each block takes to execute in milliseconds. The latency was measured on an NVIDIA A100 GPU over the entire dataset, recording the latency of each block individually, repeating each sample 100 times and averaging the results. For the detection head, the baseline exit was considered and post-processing such as non-maximum suppression is not considered. The cumulative latency is the time it takes to reach the i-th block from the first block. We note that for \gelantransposed{} this is not the optimal execution order. As previously noted, one can clearly observe that the GELAN block is the most computationally expensive part of the network, while the upsampling and concatenation layers are very cheap in comparison.

In \cref{tab:ap50-90} we show the AP$_{50:90}$ performance of the tiny AnytimeYOLO variants and in \cref{tab:ap50-90-m} we show the AP$_{50:90}$ performance of the medium AnytimeYOLO variants.

\begin{figure*}
    \begin{subfigure}[b]{0.33\textwidth}
   \centering
   \includegraphics[width=0.95\columnwidth]{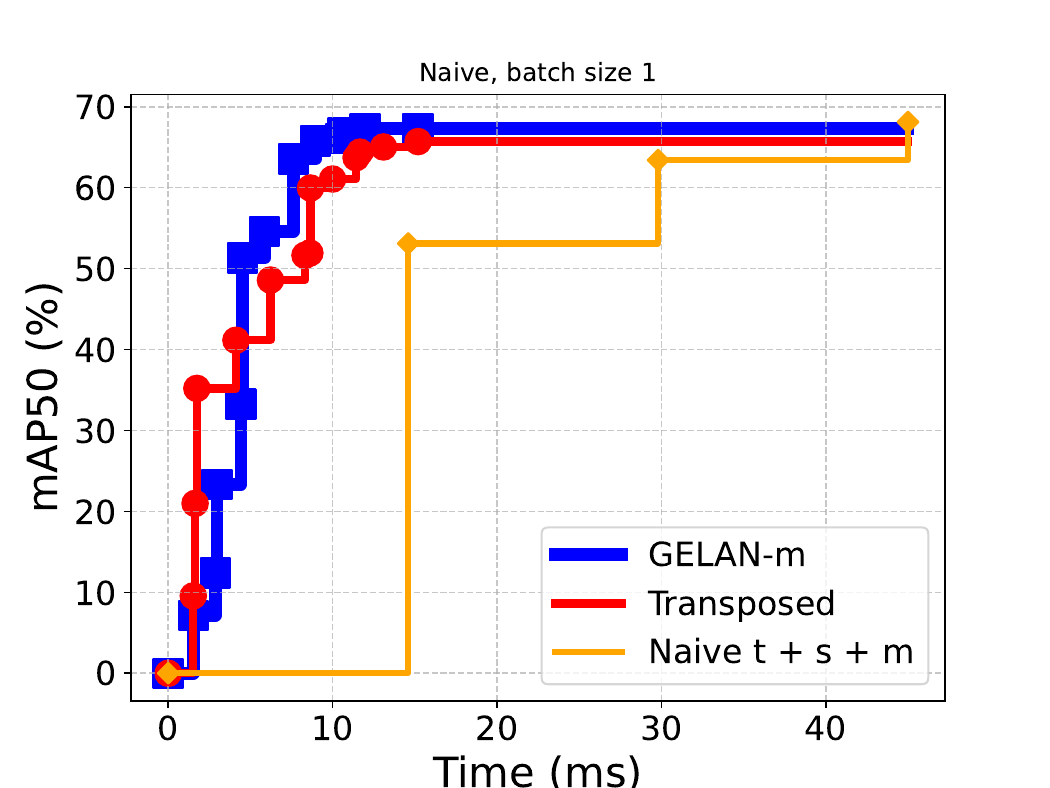}
   \caption{Batch size 1}
   \label{fig:naive-batch-size-1}
   \end{subfigure}
   \begin{subfigure}[b]{0.33\textwidth}
   \centering
   \includegraphics[width=0.95\columnwidth]{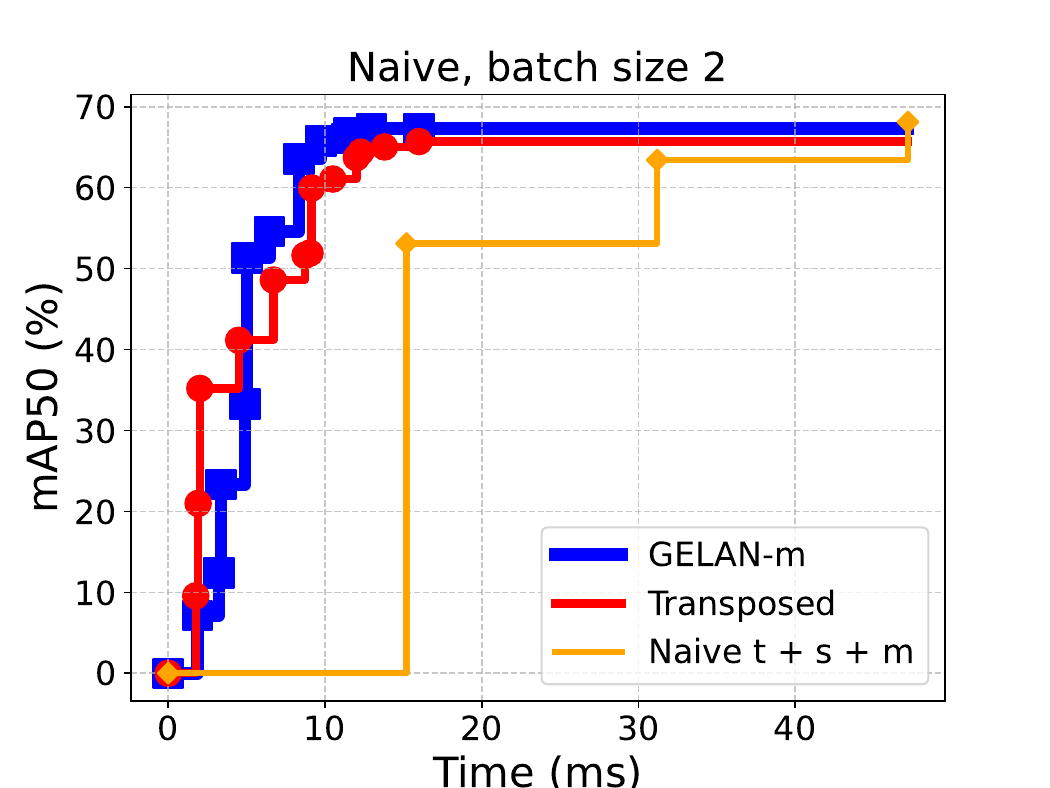}
   \caption{Batch size 2}
   \label{fig:naive-batch-size-2}
   \end{subfigure}
   \begin{subfigure}[b]{0.33\textwidth}
   \centering
   \includegraphics[width=0.95\columnwidth]{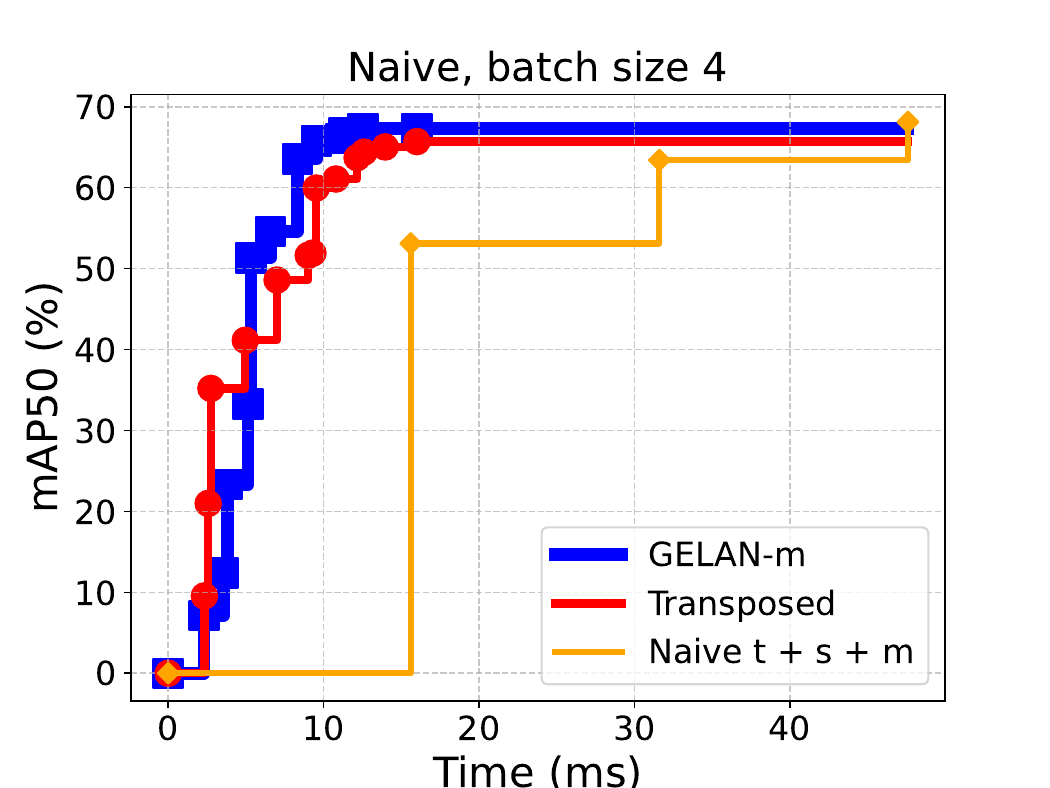}
   \caption{Batch size 4}
   \label{fig:naive-batch-size-4}
   \end{subfigure}
   \begin{subfigure}[b]{0.33\textwidth}
   \centering
   \includegraphics[width=0.95\columnwidth]{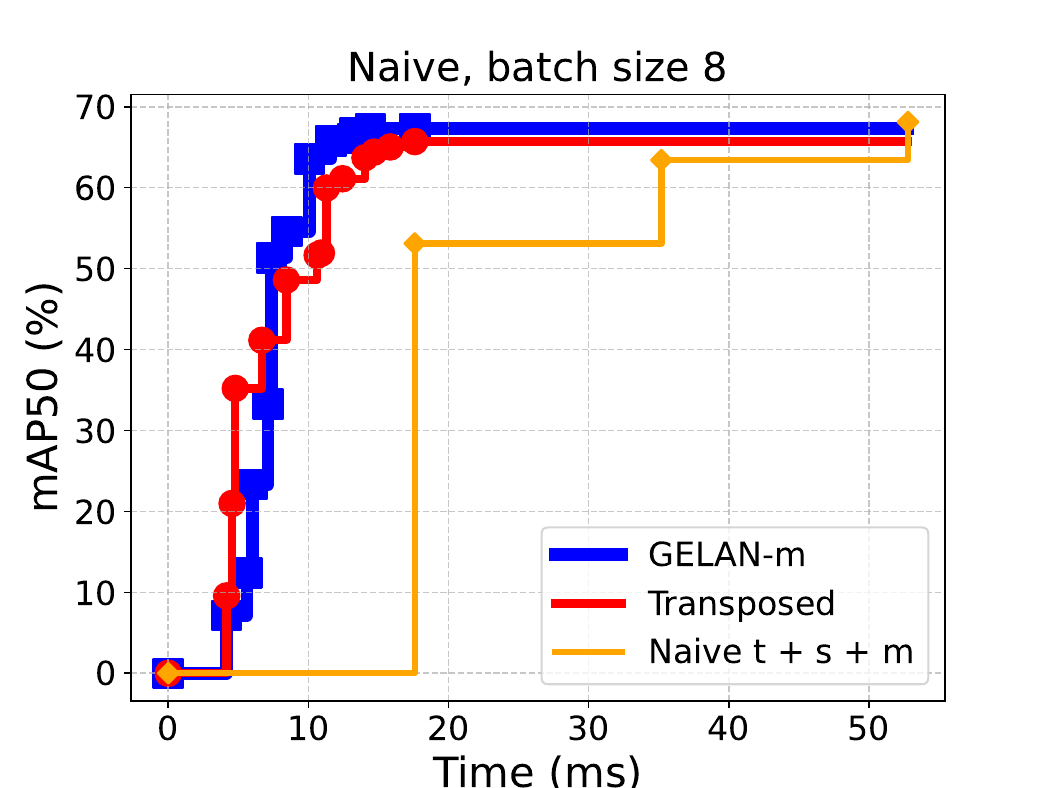}
   \caption{Batch size 8}
   \label{fig:naive-batch-size-8}
   \end{subfigure}
   \begin{subfigure}[b]{0.33\textwidth}
   \centering
   \includegraphics[width=0.95\columnwidth]{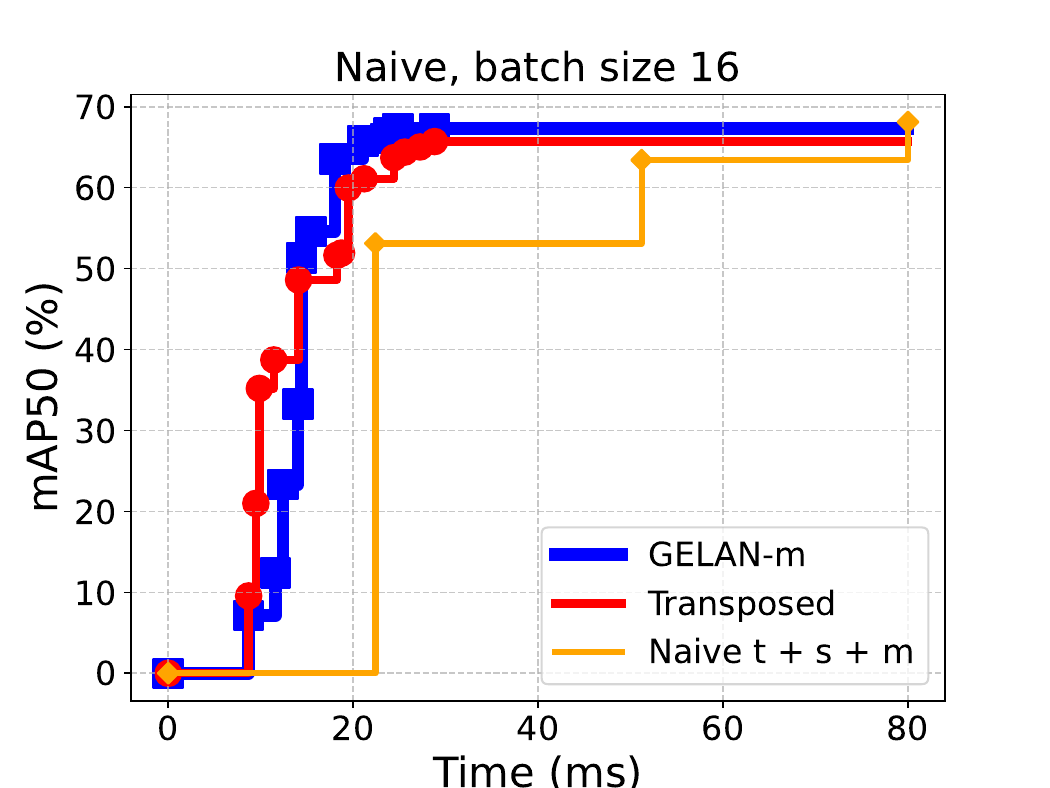}
   \caption{Batch size 16}
   \label{fig:naive-batch-size-16}
   \end{subfigure}
   \begin{subfigure}[b]{0.33\textwidth}
   \centering
   \includegraphics[width=0.95\columnwidth]{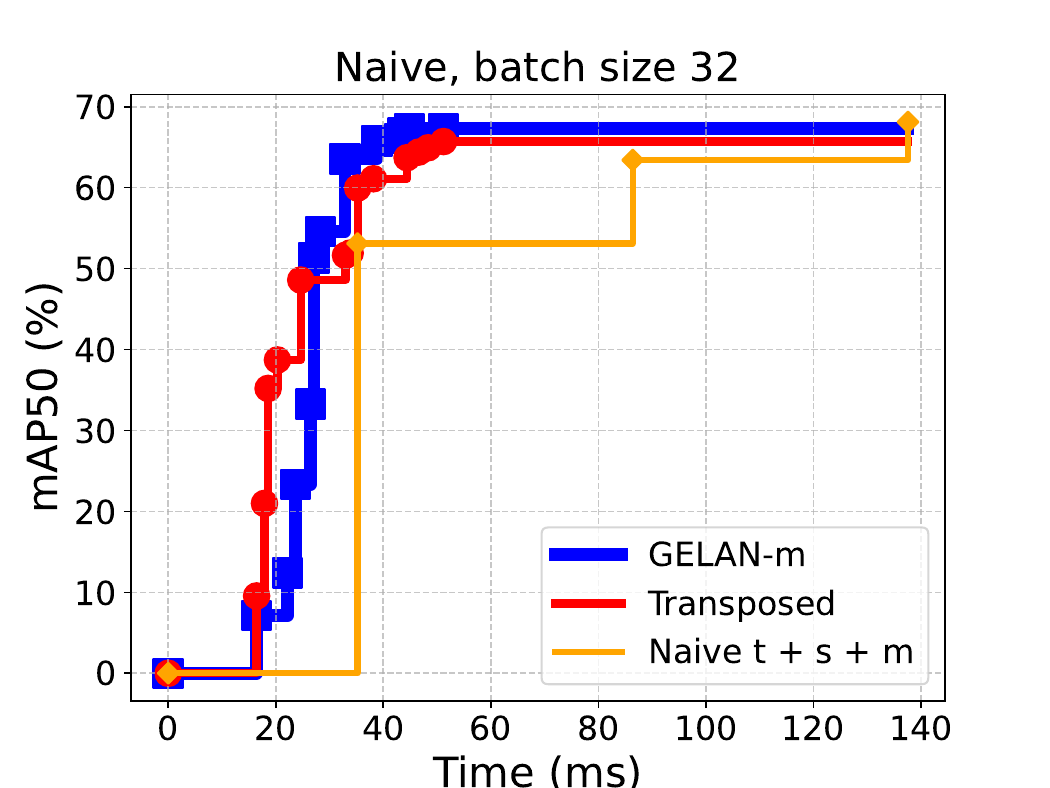}
   \caption{Batch size 32}
   \label{fig:naive-batch-size-32}
   \end{subfigure}
   \begin{subfigure}[b]{0.33\textwidth}
   \centering
   \includegraphics[width=0.95\columnwidth]{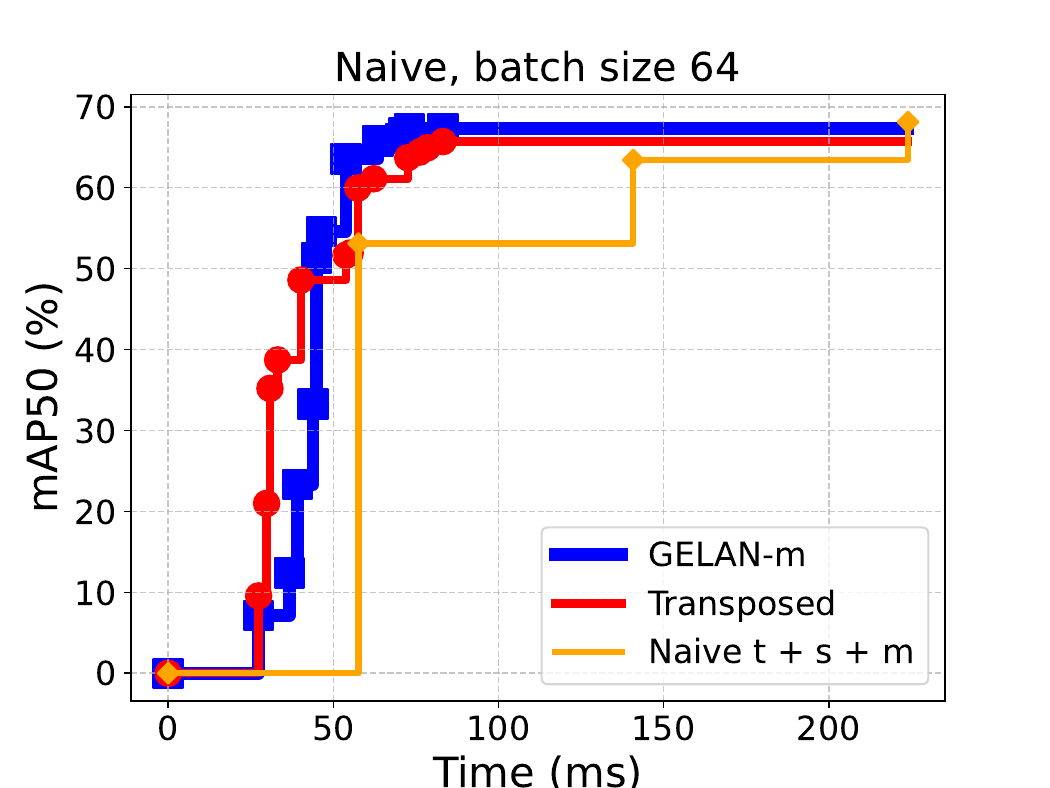}
   \caption{Batch size 64}
   \label{fig:naive-batch-size-64}
   \end{subfigure}
   \begin{subfigure}[b]{0.33\textwidth}
   \centering
   \includegraphics[width=0.95\columnwidth]{figures/full_plots/naive_batch_size_128.pdf}
   \caption{Batch size 128}
   \label{fig:naive-batch-size-128}
   \end{subfigure}
   \caption{Comparison of medium AnytimeYOLO to running YOLOv9-t, YOLOv9-s, and YOLOv9-m sequentially.}
   \label{fig:batch-size-comparison}
   \end{figure*}

In \cref{fig:batch-size-comparison} we compare the na\"ive variant of running tiny, small and then medium YOLO sequentially and medium AnytimeYOLO with 15 exits at different batch sizes. The naive variant performs comparatively much worse at lower batch sizes, as the GPU isn't fully saturated, so the tiny variant is not able to fully utilize the GPU and is therefore closer in speed to the medium variant. For higher batch sizes, the naive variant performs better as the tiny variant gets quicker comparatively to the medium variant and AnytimeYOLO needs a longer time to the first exit. However, it still performs noticeably worse than AnytimeYOLO while also providing much less granularity.

\section{Preliminary experiments}
\label{sec:suppl-preliminary-experiments}
We provide the results of our preliminary experiments, investigating the impact of the number of sub-exits on the performance of the final exit, the impact of joint training and sampling, and the impact of training for 400 epochs. These preliminary experiments were condutected for the tiny variant of AnytimeYOLO.

We conducted preliminary experiments to investigate the impact of the number of sub-exits on the performance of the final exit. Following the progressive shrinking scheme, we added sub-exits starting from the end of the \gelant{} network. Three sub-exits corresponds to the initial baseline, while fifteen sub-exits corresponds to the network with the highest granularity. The results are listed in~\cref{tab:head-count}. When adding sub-exits in this manner, we can observe that the performance of the final exit decreases continuously as the number of sub-exits increases.

As mentioned in~\cref{sec:suppl-implementation-details}, for our final experiment for the 9 sub-exit variant, we opted to spread the sub-exits evenly across the network, placing three in the beginning, three in the middle and three in the end. The performance of the final exit in this case is 49.1\% AP$_{50}$ (AP$_{50:90}$ 34.98\%), somewhat worse than even the 15 sub-exit variant. The relationship between sub-exit count and final exit performance is therefore not straightforward, depending also on the placement of the sub-exits. As we consider the higher granularity provided by spreading the exits evenly to be more beneficial, we focused on the configuration with 9 and 15 sub-exits for our final experiments.

To compare joint training and sampling, we trained a \gelantransposed{} variant with 17 sub-exits and no pre-training using both joint training and sampling. The variant trained using joint training achieved an accuracy of 47.0\% AP$_{50}$ (AP$_{50:90}$ 33.3\%), while the variant trained using sampling achieved a slightly better accuracy of 47.3\% AP$_50$ (AP$_{50:90}$ 33.6\%).  We note that the model has 17 sub-exits. In this initial experiment, we inserted two additional GELAN blocks, which slightly improved performance and anytime quality. However, this also caused approximately a 20\% increase in latency. As a result, we discarded these blocks after the initial experiments. 

Since, as previously mentioned, progressive shrinking might be impacted by the exits being spread around, we also considered a variant of \gelant{} with 9 sub-exits and no pre-training for both joint training and sampling. The former achieved 49.3\% AP$_{50}$ (AP$_{50:90}$ 35.2\%, respectively), while the latter achieved 49.1\% AP$_{50}$ (AP$_{50:90}$ 35.2\%, respectively), a slight decrease in accuracy when using sampling.

As the difference between the joint training and sampling approach are almost negligible, we chose sampling due to its lower memory and computational requirements.

We also considered training with different epochs of pre-training. However, as shown in~\cref{tab:400-epochs}, a longer pre-training period (400 epochs) can result in worse performance of the final exit. That is, the performance of the final exit is slightly worse than for 200 epochs of pre-training, though it is better than no pre-training. Therefore, we only considered no pre-training and 200 epochs of pre-training in our final experiments. We note that for \gelantransposed{} a variant was used with two additional GELAN blocks, i.e., with 17 total sub-exits.

\section{Deployment measurements}
\label{sec:suppl-deployment-measurements}
We measured the latency of \gelant{} with 15 sub-exits when deployed using TorchScript and TensorRT. The results are shown in~\cref{tab:deployment-latency}. The latency was measured on an NVIDIA RTX 3090 GPU using 16-bit precision. The baseline latency is the latency of the network without any exits. We also measured the latency when each block of the network was compiled individually (denoted as Chunked compilation in~\cref{tab:deployment-latency}).
Soft anytime is the latency when synchronizing with the host after each block so that early termination is possible. For hard anytime inference, exits under consideration must be computed when encountered along the path. We note that exits here refers not to individual sub-exits but to a tuple of sub-exits, e.g., in the case of three exits, (4, 5, 7), (9, 11, 14), (15, 18, 21) were chosen.

The results show that soft anytime inference has a much smaller impact on latency for TorchScript (around 3\%) than for TensorRT (around 82\%), the need for separate compilation playing a slightly larger role than the need to synchronize with the host. %
Hard anytime inference has a much larger impact on latency. Even with a small amount of exits. For TorchScript the slowdown is around 40\% for 3 exits, while for TensorRT it is significantly greater, at approximately 2.63 times. As the number of exits increases, the slowdown increases as well.

\begin{table}
    \centering
    \begin{tabular}{rlll}
        Index & Module & Latency & Cum. Latency \\ \hline
        0 & Conv & 0.05 ms & 0.05 ms \\
        1 & Conv & 0.04 ms & 0.09 ms \\
        2 & ELAN1 & 0.24 ms & 0.33 ms \\
        3 & Down & 0.07 ms & 0.40 ms \\
        4 & CSP-ELAN & 1.27 ms & 1.66 ms \\
        5 & Down & 0.07 ms & 1.74 ms \\
        6 & CSP-ELAN & 1.29 ms & 3.03 ms \\
        7 & Down & 0.07 ms & 3.10 ms \\
        8 & CSP-ELAN & 1.32 ms & 4.42 ms \\
        9 & SPP-ELAN & 0.18 ms & 4.60 ms \\
        10 & Up & 0.01 ms & 4.61 ms \\
        11 & Concat & 0.01 ms & 4.62 ms \\
        12 & CSP-ELAN & 1.29 ms & 5.91 ms \\
        13 & Up & 0.01 ms & 5.92 ms \\
        14 & Concat & 0.01 ms & 5.93 ms \\
        15 & CSP-ELAN & 1.27 ms & 7.21 ms \\
        16 & Down & 0.07 ms & 7.28 ms \\
        17 & Concat & 0.01 ms & 7.29 ms \\
        18 & CSP-ELAN & 1.30 ms & 8.59 ms \\
        19 & Down & 0.07 ms & 8.66 ms \\
        20 & Concat & 0.01 ms & 8.67 ms \\
        21 & CSP-ELAN & 1.32 ms & 9.99 ms \\
        22 & Detect & 1.36 ms & 11.35 ms
    \end{tabular}
    \caption{Latencies for \gelant{}.}
    \label{tab:gelant-latency}
\end{table}

\begin{table}
    \centering
\begin{tabular}{rlll}
       Index & Module & Latency   & Cum. Latency   \\
    \hline
           0 & Conv & 0.05 ms   & 0.05 ms              \\
           1 & Conv & 0.04 ms   & 0.09 ms              \\
           2 & ELAN1 & 0.23 ms   & 0.32 ms              \\
           3 & Down & 0.07 ms   & 0.40 ms              \\
           4 & Down & 0.07 ms   & 0.47 ms              \\
           5 & Down & 0.07 ms   & 0.54 ms              \\
           6 & CSP-ELAN & 1.30 ms   & 1.84 ms              \\
           7 & CSP-ELAN & 1.28 ms   & 3.12 ms              \\
           8 & CSP-ELAN & 1.26 ms   & 4.38 ms              \\
           9 & SPP-ELAN & 0.18 ms   & 4.56 ms              \\
          10 & Up & 0.01 ms   & 4.57 ms              \\
          11 & Concat & 0.01 ms   & 4.58 ms              \\
          12 & Up & 0.01 ms   & 4.59 ms              \\
          13 & Concat & 0.01 ms   & 4.60 ms              \\
          14 & CSP-ELAN & 1.28 ms   & 5.88 ms              \\
          15 & CSP-ELAN & 1.26 ms   & 7.14 ms              \\
          16 & Down & 0.07 ms   & 7.21 ms              \\
          17 & Concat & 0.01 ms   & 7.22 ms              \\
          18 & Down & 0.07 ms   & 7.29 ms              \\
          19 & Concat & 0.01 ms   & 7.30 ms              \\
          20 & CSP-ELAN & 1.31 ms   & 8.61 ms              \\
          21 & CSP-ELAN & 1.28 ms   & 9.89 ms              \\
          22 & Detect & 1.36 ms   & 11.24 ms
\end{tabular}
\caption{Latencies for \gelantransposed{}.}
\label{tab:gelantransposed-latency}
\end{table}

\begin{table}
    \centering
    \begin{tabular}{rlll}
    Index & Module & Latency & Cum. Latency \\ \hline
    0 & Conv & 0.05 ms & 0.05 ms \\
    1 & Conv & 0.06 ms & 0.11 ms \\
    2 & ELAN1 & 0.84 ms & 0.95 ms \\
    3 & Down & 0.09 ms & 1.04 ms \\
    4 & CSP-ELAN & 0.90 ms & 1.94 ms \\
    5 & Down & 0.08 ms & 2.02 ms \\
    6 & CSP-ELAN & 0.97 ms & 2.99 ms \\
    7 & Down & 0.07 ms & 3.07 ms \\
    8 & CSP-ELAN & 0.90 ms & 3.97 ms \\
    9 & SPP-ELAN & 0.18 ms & 4.15 ms \\
    10 & Up & 0.01 ms & 4.16 ms \\
    11 & Concat & 0.01 ms & 4.17 ms \\
    12 & CSP-ELAN & 0.97 ms & 5.14 ms \\
    13 & Up & 0.01 ms & 5.15 ms \\
    14 & Concat & 0.01 ms & 5.16 ms \\
    15 & CSP-ELAN & 0.91 ms & 6.07 ms \\
    16 & Down & 0.07 ms & 6.14 ms \\
    17 & Concat & 0.01 ms & 6.15 ms \\
    18 & CSP-ELAN & 0.97 ms & 7.12 ms \\
    19 & Down & 0.07 ms & 7.19 ms \\
    20 & Concat & 0.01 ms & 7.20 ms \\
    21 & CSP-ELAN & 0.91 ms & 8.11 ms \\
    22 & Detect & 1.37 ms & 9.48 ms
    \end{tabular}
    \caption{Latencies for \gelanm{}.}
    \label{tab:gelanm-latency}
    \end{table}

    \begin{table}
        \centering
        \begin{tabular}{rlll}
        Index & Module & Latency & Cum. Latency \\
        \hline
        0 & Conv & 0.05 ms & 0.05 ms \\
        1 & Conv & 0.06 ms & 0.12 ms \\
        2 & ELAN1 & 0.92 ms & 1.04 ms \\
        3 & Down & 0.09 ms & 1.13 ms \\
        4 & Down & 0.08 ms & 1.21 ms \\
        5 & Down & 0.08 ms & 1.29 ms \\
        6 & CSP-ELAN & 1.55 ms & 2.84 ms \\
        7 & CSP-ELAN & 1.76 ms & 4.61 ms \\
        8 & CSP-ELAN & 1.56 ms & 6.16 ms \\
        9 & SPP-ELAN & 0.20 ms & 6.36 ms \\
        10 & Up & 0.01 ms & 6.37 ms \\
        11 & Concat & 0.01 ms & 6.38 ms \\
        12 & Up & 0.01 ms & 6.39 ms \\
        13 & Concat & 0.02 ms & 6.41 ms \\
        14 & CSP-ELAN & 1.06 ms & 7.47 ms \\
        15 & CSP-ELAN & 1.00 ms & 8.47 ms \\
        16 & Down & 0.08 ms & 8.55 ms \\
        17 & Concat & 0.01 ms & 8.56 ms \\
        18 & Down & 0.08 ms & 8.64 ms \\
        19 & Concat & 0.01 ms & 8.65 ms \\
        20 & CSP-ELAN & 1.56 ms & 10.21 ms \\
        21 & CSP-ELAN & 1.06 ms & 11.27 ms \\
        22 & Detect & 1.51 ms & 12.78 ms
        \end{tabular}
        \caption{Latencies for \gelanmtransposed{}.}
        \label{tab:gelanmtransposed-latency}
        \end{table}
\begin{table}
    \centering
    \scalebox{0.8}{
    \begin{tabular}{l|c|c|c|c|l}
        \textbf{Model} & \textbf{Pretrain} & \textbf{Exits} & \textbf{$Q_{AP_{50:90}}$} & \textbf{$Q_{AP_{50:90}SE}$} & $\text{AP}_{50:90}$ \\ \hline \hline
        YOLOv9-t & - & - & - & - & 38.30 \\
\gelant & - & - & - & - & 36.90 \\
\gelantransposed & - & - & - & - & 35.08 \\
\hline
\gelant & 0 & 9 & 22.44 & 3.53 & 34.98 \\
\gelant & 0 & 15 & 23.3 & 3.35 & 35.56 \\
\gelant & 200 & 9 & 22.29 & 3.64 & 36.07 \\
\gelant & 200 & 15 & 23.10 & 3.48 & 36.31 \\
\gelantransposed & 0 & 9 & 22.44 & 2.88 & 33.27 \\
\gelantransposed & 0 & 15 & 23.59 & 2.61 & 33.33 \\
\gelantransposed & 200 & 9 & 22.25 & 3.01 & 34.13 \\
\gelantransposed & 200 & 15 & 22.89 & 2.84 & 33.79 \\
\end{tabular}
}
\caption{AP$_{50:90}$ performance of AnytimeYOLO \gelant{} and \gelantransposed variants.}
\label{tab:ap50-90}
\end{table}

\begin{table}
    \centering
    \scalebox{0.8}{
    \begin{tabular}{l|c|c|c|c|l}
        \textbf{Model} & \textbf{Pretrain} & \textbf{Exits} & \textbf{$Q_{AP_{50:90}}$} & \textbf{$Q_{AP_{50:90}SE}$} & $\text{AP}_{50:90}$ \\ \hline \hline
        YOLOv9-m & - & - & - & - & 51.40 \\
\gelanm & - & - & - & - &  50.96 \\
\gelanmtransposed & - & - & - & - & 49.85 \\
\hline
\gelanm & 0 & 9 & 34.22 & 6.26 & 50.14 \\
\gelanm & 0 & 15 & 34.05 & 6.43 & 50.73 \\
\gelanm & 200 & 9 & 33.84 & 6.41 & 50.90 \\
\gelanm & 200 & 15 & 34.84 & 6.12 & 50.91 \\
\gelanmtransposed & 0 & 9 & 31.59 & 5.72 & 49.08 \\
\gelanmtransposed & 0 & 17 & 31.81 & 5.87 & 49.19 \\
\gelanmtransposed & 200 & 9 & 30.44 & 6.13 & 49.58 \\
\gelanmtransposed & 200 & 17 & 32.46 & 5.68 & 49.33 \\
\end{tabular}
}
\caption{AP$_{50:90}$ performance of AnytimeYOLO \gelanm{} and \gelanmtransposed{} variants.}
\label{tab:ap50-90-m}
\end{table}

\begin{table}
\centering
\scalebox{0.8}{
\begin{tabular}{l|c|c}
\textbf{Sub-exits} & \textbf{AP$_{50}$} & \textbf{AP$_{50:90}$} \\ \hline \hline
3 & 51.34 & 36.90 \\
5 & 50.97 & 36.60 \\
7 & 50.60 & 36.18 \\
9 & 50.36 & 35.97 \\
12 & 50.23 & 35.94 \\
15 & 49.84 & 35.56    
\end{tabular}
}
\caption{\gelant{} (no pre-training) with different sub-exits counts.}
\label{tab:head-count}
\end{table}

\begin{table}
\centering
\scalebox{0.8}{
\begin{tabular}{l|c|c|c|c}
\textbf{Model} & \textbf{Pre-training} & \textbf{Sub-exits} & \textbf{AP$_{50}$} & \textbf{AP$_{50:90}$} \\ \hline \hline
\gelant & 0 & 9 & 49.10 & 34.98 \\
\gelant & 0 & 15 & 49.84 & 35.56 \\
\gelant & 200 & 9 & 50.61 & 36.27 \\
\gelant & 200 & 15 & 50.49 & 36.26 \\
\gelant & 400 & 9 & 50.04 & 35.82 \\
\gelant & 400 & 15 & 49.92 & 35.65 \\

\gelantransposed & 0 & 9 & 48.42 & 34.56 \\
\gelantransposed & 0 & 17 & 47.26 & 33.58 \\
\gelantransposed & 200 & 9 & 48.59 & 34.47\\
\gelantransposed & 200 & 17 & 48.86 & 34.80\\
\gelantransposed & 400 & 9 & 48.55 & 34.42 \\
\gelantransposed & 400 & 17 & 48.74 & 34.53 

\end{tabular}
}
\caption{AP$_{50:90}$ performance of AnytimeYOLO variants with 400 epochs of pre-training.}
\label{tab:400-epochs}
\end{table}

\begin{table}
    \centering
    \begin{tabular}{l|c|c}
    \textbf{Method} & \textbf{Latency (ms)} & \textbf{Slowdown} \\ \hline \hline
    \multicolumn{3}{c}{\textbf{TorchScript}} \\ \hline
    Baseline & 7.56 & 1.00 \\
    Chunked tracing & 7.63 & 1.01 \\
    Soft Anytime & 7.83 & 1.03 \\
    Hard Anytime, 3 exits & 10.56 & 1.40 \\
    Hard Anytime, 9 exits & 16.23 & 2.14 \\
    Hard Anytime, 15 exits & 23.77 & 3.14 \\ \hline
    \multicolumn{3}{c}{\textbf{TensorRT}} \\ \hline
    Baseline & 3.03 & 1.00 \\
    Chunked compilation & 4.84 & 1.48 \\
    Soft Anytime & 5.53 & 1.82 \\
    Hard Anytime, 3 exits & 7.97 & 2.63 \\
    Hard Anytime, 9 exits & 11.07 & 3.65 \\
    Hard Anytime, 15 exits & 15.80 & 5.21 \\ \hline
    \end{tabular}
    \caption{Latency measurements for \gelant{} deployment using TorchScript and TensorRT on an NVIDIA RTX 3090.}
\label{tab:deployment-latency}
\end{table}

\end{document}